\documentclass{article}

\usepackage[utf8]{inputenc} 
\usepackage[T1]{fontenc}    
\usepackage{url}            
\usepackage{booktabs}       
\usepackage{amsfonts}       
\usepackage{nicefrac}       
\usepackage{microtype}      
\usepackage{pifont}

\usepackage{xspace}
\usepackage{color}
\usepackage{enumitem}
\usepackage{graphicx}
\usepackage{wrapfig}
\usepackage{mathtools}
\usepackage{multirow}
\usepackage[dvipsnames]{xcolor}
\usepackage[labelfont=bf,textfont=md]{caption}
\usepackage{subcaption}
\usepackage{fullwidth}
\usepackage{etoolbox}
\usepackage{enumitem}

\usepackage[nohyperref,accepted]{icml2020}

\usepackage{macros}

\icmltitlerunning{The Non-IID Data Quagmire of Decentralized Machine Learning}

\patchcmd{\quote}{\rightmargin}{\leftmargin 1em \rightmargin}{}{}

\begin{document}

\twocolumn[
\icmltitle{The Non-IID Data Quagmire of Decentralized Machine Learning}




\begin{icmlauthorlist}
\icmlauthor{Kevin Hsieh}{ms,cmu}
\icmlauthor{Amar Phanishayee}{ms}
\icmlauthor{Onur Mutlu}{eth,cmu}
\icmlauthor{Phillip B. Gibbons}{cmu}
\end{icmlauthorlist}

\icmlaffiliation{ms}{Microsoft Research}
\icmlaffiliation{eth}{ETH Z{\"u}rich}
\icmlaffiliation{cmu}{Carnegie Mellon University}

\icmlcorrespondingauthor{Kevin Hsieh}{kevin.hsieh@microsoft.com}
\icmlcorrespondingauthor{Phillip B. Gibbons}{gibbons@cs.cmu.edu}

\icmlkeywords{decentralized learning, federated learning, geo-distributed learning, non-iid data}

\vskip 0.3in
]

\printAffiliationsAndNotice{}

\begin{abstract}
Many large-scale machine learning (ML) applications need to perform \emph{decentralized} learning over datasets generated at different devices and locations. Such datasets pose a significant challenge to decentralized learning because their different contexts result in significant data distribution skew across devices/locations. In this paper, we take a step toward better understanding this challenge by presenting a detailed experimental study of decentralized DNN training on a common type of data skew: skewed distribution of data labels across \kv{devices/locations}. Our study shows that: (i) skewed data labels are a fundamental and pervasive problem for decentralized learning, causing significant accuracy loss across many ML applications, DNN models, training datasets, and decentralized learning algorithms; (ii) the problem is particularly challenging for DNN models with batch normalization; and (iii) the degree of data skew is a key determinant of the difficulty of the problem. Based on these findings, we present \name, a system-level approach that adapts the communication frequency of decentralized learning algorithms to the (skew-induced) accuracy loss between data partitions. We also show that group normalization can recover much of the accuracy loss of batch normalization.
\end{abstract}

\section{Introduction}
\label{sec:intro}
\vspace{-0.05in}
The advancement of machine learning (ML) is heavily dependent on the processing of massive amounts of data.
The most timely and relevant data are often generated at different devices all over the world, e.g., data collected by mobile phones and video cameras.
Because of communication and privacy constraints, gathering all \kv{such} data for centralized processing can be impractical/infeasible.
For example, moving raw data across national borders is subject to data sovereignty law constraints (e.g., GDPR\kv{~\cite{regulation2016regulation}}). 
Similar constraints apply to centralizing private data from phones. 

These constraints motivate the need for ML training over widely distributed data ({\bf\emph{decentralized learning}}).
For example, \emph{geo-distributed learning}~\cite{DBLP:conf/nsdi/HsiehHVKGGM17} trains a global ML model over data spread across geo-distributed data centers. 
Similarly, \emph{federated learning}~\cite{DBLP:conf/aistats/McMahanMRHA17} trains a centralized model over data from a large number of mobile devices.
Federated learning has been an important topic both in academia (140+ papers in 2019) and industry (500+ million installations on Android devices).

\textbf{Key Challenges in Decentralized Learning.} There are two key challenges in decentralized learning. 
First, training a model over decentralized data using traditional training approaches (i.e., those designed for centralized data, often using a bulk synchronous parallel (BSP) approach~\cite{DBLP:journals/cacm/Valiant90}) requires massive \kv{amounts of} communication.  
\kv{Doing so} drastically slows down the training process because the communication is bottlenecked by the limited wide-area or mobile network bandwidth~\cite{DBLP:conf/nsdi/HsiehHVKGGM17, DBLP:conf/aistats/McMahanMRHA17}.
Second, decentralized data \kv{is} typically generated at different contexts, which can lead to significant differences in the \emph{distribution} of data across data partitions.
For example, facial images collected by cameras \kv{would} reflect the demographics of each camera's location, and images of kangaroos \kv{can} be collected only from cameras in Australia or zoos.
Unfortunately, existing decentralized learning algorithms (e.g., \cite{DBLP:conf/nsdi/HsiehHVKGGM17, DBLP:conf/aistats/McMahanMRHA17, DBLP:conf/nips/SmithCST17, DBLP:conf/ICLR/LinHMWD18, DBLP:conf/icml/TangLYZL18}) mostly focus on reducing communication, as they either (i) assume the data partitions are independent and identically distributed (IID) or (ii) conduct only very limited \kv{studies} on non-IID data partitions. This leaves a key question mostly unanswered: \emph{What happens to ML applications and decentralized learning algorithms when their data partitions are not IID?}

\textbf{Our Goal and Key Findings.} 
We aim to take a step to further the understanding of the above key question. 
In this work, we focus on a common type of non-IID data, widely used in prior work (e.g., \cite{DBLP:conf/aistats/McMahanMRHA17, DBLP:conf/icml/TangLYZL18, DBLP:journals/corr/abs-1806-00582}): skewed distribution of data labels across \kv{devices/locations}.
Such \textit{skewed label partitions} arise frequently in the real world (see \xref{subsec:dataset} for examples).
Our study covers various DNN applications, DNNs, training datasets, decentralized learning algorithms, and degrees of label skew.
Our study reveals three key findings:

\squishlist
\item Training over skewed label partitions is a fundamental and pervasive problem for decentralized learning. 
Three decentralized learning algorithms~\cite{DBLP:conf/nsdi/HsiehHVKGGM17, DBLP:conf/aistats/McMahanMRHA17, DBLP:conf/ICLR/LinHMWD18} suffer from major model quality loss when run to convergence on skewed label partitions, across the applications, models, and training datasets in our study.
\item DNNs with \emph{batch normalization}~\cite{DBLP:conf/icml/IoffeS15} are particularly vulnerable to skewed label partitions, suffering significant model quality loss even under BSP, the most communication-heavy approach.
\item The degree of skew is a key determinant of the difficulty level of the problem.
\squishlistend
These findings reveal that non-IID data is an important yet heavily understudied challenge in decentralized learning, worthy of extensive study.
To facilitate further study on skewed label partitions, we release a real-world, geo-tagged dataset of common mammals on Flickr~\cite{flickr}, \kv{which is openly available at \textcolor{Blue}{\url{https://doi.org/10.5281/zenodo.3676081}}} (\xref{subsec:dataset}).

\textbf{Solutions.}
As two initial steps towards addressing the vast challenge of non-IID data, we first 
show that among the many proposed alternatives to batch normalization, \emph{group normalization}~\cite{DBLP:conf/eccv/WuH18} avoids the skew-induced accuracy loss of batch normalization under BSP.
With this fix, all models in our study achieve high accuracy on skewed label partitions under (communication-heavy) BSP, and the problem can be viewed as a trade-off between accuracy and the amount of communication.
Intuitively, there is a tug-of-war among different data partitions, with each partition pulling the model to reflect its data, and only close communication, tuned to the skew-induced accuracy loss, can save the overall model accuracy of the algorithms in our study.
Accordingly, we present \name, which periodically sends local models to remote data partitions and compares the model performance (e.g., validation accuracy) between local and remote partitions. 
Based on the accuracy loss, {\name} adjusts the amount of communication among data partitions by controlling how \emph{relaxed} the decentralized learning algorithms should be, such as controlling the threshold that determines which parameters are worthy of communication.
Thus, {\name} can seamlessly integrate with decentralized learning algorithms that provide such communication control.
Our experimental results show that \name's adaptive approach automatically reduces communication by 9.6$\times$ (under high skew) to 34.1$\times$ (under mild skew) while retaining the accuracy of BSP.

\textbf{Contributions.} We make the following contributions. 
First, we conduct a detailed empirical study on the problem of skewed label partitions. 
We show that this problem is a fundamental and pervasive challenge for decentralized learning.
Second, we build and release a large real-world dataset to facilitate future study on this challenge.
Third, we make a new observation showing that this challenge is particularly problematic for DNNs with batch normalization, even under BSP.
We discuss the root cause of this problem and we find that it can be addressed by using an alternative normalization technique.
Fourth, we show that the difficulty level of this problem varies with the data skew.
Finally, we design and evaluate \name, a system-level approach that adapts the communication frequency \kv{among data partitions} to reflect the skewness in the data, seeking to maximize communication savings while preserving model accuracy.


\section{Background and Motivation}
\label{sec:background}

We first provide background on popular decentralized learning algorithms (\xref{subsec:decentral}).
We then highlight a real-world example of skewed label partitions: geographical distribution of \pg{mammal} pictures on Flickr, among other examples (\xref{subsec:dataset}).

\subsection{Decentralized Learning}
\label{subsec:decentral}

In a decentralized learning setting, we aim to train \pg{an} ML model $w$ based on all the training data samples $(x_i, y_i)$ that are generated and stored in one of the $K$ partitions (denoted as $P_k$). The goal of the training is to fit $w$ to all data samples. Typically, most decentralized learning algorithms assume the data samples are independent and identically distributed (IID) among different $P_k$, and we refer to such a setting as the \emph{IID setting}. Conversely, we call it the \emph{Non-IID setting} if such an assumption does not hold. 

We evaluate three popular decentralized learning algorithms to see how they perform on different applications over the IID and Non-IID settings, using skewed label partitions.
These algorithms can be used with a variety of stochastic gradient descent (SGD)\kv{~\cite{robbins1951stochastic}} approaches, and aim to reduce communication, either among data partitions ($P_k$) or between the data partitions and a centralized server. 

\squishlist
 \item \gaia~\cite{DBLP:conf/nsdi/HsiehHVKGGM17}, a geo-distributed learning algorithm that dynamically eliminates insignificant communication among data partitions. 
 Each partition $P_k$ accumulates updates $\Delta w_j$ to each model weight $w_j$ locally, and communicates $\Delta w_j$ to all other data partitions only when its relative magnitude exceeds a predefined threshold (Algorithm~\ref{algo:gaia} in Appendix~\ref{appendix:decentral_algo}\footnote{All Appendices are in the supplemental material.}).
 
 \item \fedavg~\cite{DBLP:conf/aistats/McMahanMRHA17}, a popular algorithm for federated learning that combines local SGD on each client with model averaging.
 \kv{The algorithm} selects a subset of the partitions $P_k$ in each epoch, runs a pre-specified number of local SGD steps on each selected $P_k$, and communicates the resulting models back to a centralized server.
The server averages all these models and uses the averaged $w$ as the starting point for the next epoch. (Algorithm~\ref{algo:fedavg} in Appendix~\ref{appendix:decentral_algo}).

 \item \dgc~\cite{DBLP:conf/ICLR/LinHMWD18}, a popular algorithm that communicates only a pre-specified amount of gradients each training step, with various techniques to retain model quality such as momentum correction, gradient clipping~\cite{DBLP:conf/icml/PascanuMB13}, momentum factor masking, and warm-up training~\cite{DBLP:journals/corr/GoyalDGNWKTJH17} (Algorithm~\ref{algo:dgc} in Appendix~\ref{appendix:decentral_algo}).
\squishlistend

In addition to these decentralized learning algorithms, we show the results of using BSP~\cite{DBLP:journals/cacm/Valiant90} over the IID and Non-IID settings.
BSP is significantly slower than the above algorithms because it does not seek to reduce communication: all updates from each $P_k$ are shared among all data partitions after each training step.
As noted earlier, for decentralized learning, there is a natural tension between the amount of communication and the quality of the resulting model. \kv{Different data} distributions among the $P_k$ pull the model in different directions---more communication helps mitigate this ``tug-of-war'' so that the model well-represents all the data.
Thus, BSP, with its \pg{full communication at every step}, is used \pg{to establish a quality target for trained models}. 

\subsection{Real-World Examples of Skewed Label Partitions}
\label{subsec:dataset}

Non-IID data among devices/locations encompass many different forms. 
There can be skewed distribution of features (\pg{probability} $\mathcal{P} (x)$), labels (\pg{probability} $\mathcal{P} (y)$), or the relationship between features and labels (e.g., varying $\mathcal{P} (y|x)$ or $\mathcal{P} (x|y)$) among devices/locations~\cite{DBLP:journals/corr/abs-1912-04977} (see more discussion in Appendix~\ref{appendix:discussion}).
In this work, we focus on skewed distribution of labels ($\mathcal{P}_{P_i}(y) \not\sim \mathcal{P}_{P_j}(y)$ for different data partitions $P_i$ and $P_j$), which is also the setting considered by most prior work in this domain (e.g., \cite{DBLP:conf/aistats/McMahanMRHA17, DBLP:conf/icml/TangLYZL18, DBLP:journals/corr/abs-1806-00582}).

Skewed distribution of labels is common whenever data are generated from heterogeneous users or locations. 
For example, pedestrians and bicycles are more common in street cameras than \kv{in} highway cameras~\cite{DBLP:journals/corr/abs-1910-11089}. 
In facial recognition tasks, most individuals appear in only a few locations around the world.
Certain types of clothing (mittens, cowboy boots, kimonos, etc.) \kv{are} nearly non-existent in many parts of the world.
Similarly, certain \pg{mammals} (e.g., kangaroos) are far more likely to show up in certain locations (Australia). 
In the rest of this section, we highlight this phenomenon with a study of the geographical distribution of \pg{mammal} pictures on Flickr~\cite{flickr}. 

\textbf{Dataset Creation.} 
We start with the 48 classes in the \emph{mammal} subcategory of the 600 most common classes for bounding boxes in Open Images V4~\cite{DBLP:journals/corr/abs-1811-00982}. 
For each class label, we use Flickr's API to search for relevant pictures. 
Due to noise in Flickr search results (e.g., ``jaguar'' returns the \pg{mammal} or the car), we clean the data with a state-of-the-art DNN, PNAS~\cite{DBLP:conf/eccv/LiuZNSHLFYHM18}, which is pre-trained on ImageNet.
As we can only clean classes that exist in both Open Images and ImageNet, we end up with \pg{41} mammal classes and 736,005 total pictures.
\kv{We call the resulting dataset the \textit{\FlickrMammal} dataset} (see Appendix~\ref{appendix:dataset} for more details).

\textbf{Geographical Analysis.} We map each Flickr picture's geotag to its corresponding geographic regions based on the M49 Standard~\cite{united2019methodology}. 
As we are mostly interested in \kv{the distribution of labels ($\mathcal{P}(y)$)} among different regions, we normalize the number of samples across region (\kv{non-normalized} results are similar (Appendix~\ref{appendix:dataset})). 
Table~\ref{tbl:top_animal_continent} illustrates the top-5 classes among first-level regions (continents) and their normalized share of samples in the world. 

\begin{table}[h!]
  \centering
  \scriptsize
  \setlength{\tabcolsep}{4pt}
  \begin{tabular}{c l l l l l}
    \toprule
    \textbf{Region} & \textbf{Top 1} & \textbf{Top 2} & \textbf{Top 3} & \textbf{Top 4} & \textbf{Top 5}\\ \midrule
    \textbf{Africa} & \scell{zebra\\ (72\%)} & \scell{antelope\\ (71\%)} & \scell{lion\\(68\%)} & \scell{cheetah\\ (62\%)} & \scell{hippopotamus\\ (59\%)} \\ \midrule
    \textbf{Americas} & \scell{mule\\ (84\%)} & \scell{skunk\\ (82\%)} & \scell{armadillo\\(73\%)} & \scell{harbor seal\\ (65\%)} & \scell{squirrel\\ (61\%)} \\ \midrule
    \textbf{Asia} & \scell{panda\\ (64\%)} & \scell{hamster\\ (59\%)} & \scell{monkey\\(58\%)} & \scell{camel\\ (51\%)} & \scell{red panda\\ (42\%)} \\ \midrule
    \textbf{Europe} & \scell{lynx\\ (72\%)} & \scell{hedgehog\\ (56\%)} & \scell{sheep\\(56\%)} & \scell{deer\\ (43\%)} & \scell{otter\\ (43\%)} \\ \midrule
    \textbf{Oceania} & \scell{kangaroo\\ (92\%)} & \scell{koala\\ (92\%)} & \scell{whale\\(44\%)} & \scell{sea lion\\ (34\%)} & \scell{alpaca\\ (32\%)} \\ 
    \bottomrule \\
  \end{tabular}
  \vspace{-10pt}
  \caption{Top-5 mammals in each continent and their share of samples worldwide (e.g., 72\% of zebra images are from Africa).}
  \label{tbl:top_animal_continent}
\end{table}

\textbf{Skewed distribution of labels is a natural phenomenon.} As Table~\ref{tbl:top_animal_continent} shows, the top-5 mammals in each continent \kv{constitute} 32\%--92\% of \kv{the} normalized sample share in the world (compared to 20\% if the distribution were IID). 
As expected, the top mammals in each region reflect their population share in the world (e.g., kangaroos/koalas in Oceania and zebras/antelopes in Africa).  
Furthermore, there is \emph{no overlap} for the top-5 classes among different continents, which suggests drastically different label distributions ($\mathcal{P}(y)$) among continents.
We \kv{observe a} similar phenomenon when the analysis is done based on second-level geographical regions (Appendix~\ref{appendix:dataset}).
\kv{Our observations show} that in a decentralized learning setting, where such images would be collected and stored in their native regions,  the distribution of labels across partitions would be highly skewed. 

\section{\kv{Experimental} Setup}
\label{sec:setup}

Our study consists of three dimensions: \emph{(i)} ML applications/models, \emph{(ii)} decentralized learning algorithms, and \emph{(iii)} degree of \kv{data skew}. 
We explore all three dimensions with rigorous experimental methodologies. 
In particular, we make sure the accuracy of our trained ML models on IID data matches the reported accuracy in corresponding papers. 
\kv{All source code and settings are available at \textcolor{Blue}{\url{https://github.com/kevinhsieh/non_iid_dml}}.}

\textbf{Applications.} We evaluate different deep learning applications, DNN model structures, and training datasets:

\squishlist
 \item \appimage with four DNN models: AlexNet~\cite{DBLP:conf/nips/KrizhevskySH12}, GoogLeNet~\cite{DBLP:conf/cvpr/SzegedyLJSRAEVR15}, LeNet~\cite{lecun1998gradient}, and ResNet~\cite{DBLP:conf/cvpr/HeZRS16}.
 We use two datasets, CIFAR-10~\cite{krizhevsky2009learning} and ImageNet~\cite{ILSVRC15}.
 We use the default validation set \kv{of each of the} two datasets to \kv{quantify} the validation accuracy as our model quality metric.
 We use popular datasets in order to compare model accuracy with existing work, and we also report results with our \textit{\FlickrMammal} dataset.
 \item \appface with the center-loss face 
 model \cite{DBLP:conf/eccv/WenZL016} over the CASIA-WebFace~\cite{DBLP:journals/corr/YiLLL14a} dataset. We use verification accuracy on the LFW dataset~\cite{LFWTech} as \pg{our} model quality metric.
\squishlistend

For all applications, we tune the training parameters (e.g., learning rate, minibatch size, number of epochs, etc.) such that the baseline model (BSP in the IID setting) achieves the model quality of the \kv{corresponding} original paper. 
We then use these training parameters in all other settings. 
We further ensure that training/validation accuracy has stopped improving by the end of all our experiments. Appendix~\ref{appendix:training_parameter} lists all major training parameters in our study.

\textbf{Non-IID Data Partitions.} 
In addition to studying \textit{\FlickrMammal},
we create non-IID data partitions by partitioning datasets using the data labels, i.e., using image classes for image classification and person identities for face recognition.
We control the \emph{skewness} by controlling the fraction of data that are non-IID. 
For example, 20\% non-IID indicates 20\% of the dataset \pg{is} partitioned by labels, while the remaining 80\% \pg{is} partitioned \pg{uniformly at random}.
\xref{sec:overview} \pg{and} \xref{sec:batch_norm} focus on the 100\% non-IID setting in which \pg{the entire dataset is} partitioned \kv{using labels}, while \xref{sec:skewness} studies the effect of varying the \pg{skewness}. 
As our goal is to train a global model, the model is tested on the entire validation set.

\textbf{Hyper-Parameters Selection.} 
The algorithms we study provide the following hyper-parameters (see Appendix~\ref{appendix:decentral_algo} for details of these algorithms) to control the amount of communication (and hence the training time):

\squishlist
\item \gaia uses $T_{0}$, the initial threshold to determine if \kv{an update ($\Delta w_j$)} is significant. 
The significance threshold decreases whenever the learning rate decreases. 

\item \fedavg uses $Iter_{Local}$ to control the number of local SGD steps on each selected $P_k$.

\item \dgc uses $s$ to control the sparsity of updates (update magnitudes in top $s$ percentile are exchanged). 
Following the original paper~\cite{DBLP:conf/ICLR/LinHMWD18}, $s$ follows a warm-up schedule: 75\%, 93.75\%, 98.4375\%, 99.6\%, 99.9\%. 
We use a hyper-parameter $E_{warm}$, the number of epochs for each warm-up sparsity, to control the duration of the warm-up.
For example, if $E_{warm} = 4$, $s$ is 75\% in epochs 1--4, 93.75\% in epochs 5--8, and so on.
\squishlistend

We select a hyper-parameter $\theta$ of each decentralized learning algorithm ($T_{0}$, $Iter_{Local}$, $E_{warm}$) so that \emph{(i)} $\theta$ achieves the same model quality as BSP in the IID setting and \emph{(ii)} $\theta$ achieves similar communication savings across the three decentralized learning algorithms. 
We study the sensitivity of our findings to the choice of $\theta$ in \xref{subsec:decentral_parameter}.

\section{Non-IID Study: Results Overview}
\label{sec:overview}

This paper seeks to answer the question \kv{of} what happens to ML applications, ML models, and decentralized learning algorithms when their data label partitions are not IID.
In this section, we provide an overview of our findings, showing that
skewed label partitions cause \emph{major model quality loss}, across many applications, models, and algorithms.

\subsection{Image Classification}
\label{subsec:overview_cifar10}

\begin{figure*}[t!]
\centering  
\includegraphics[width=0.95\textwidth]{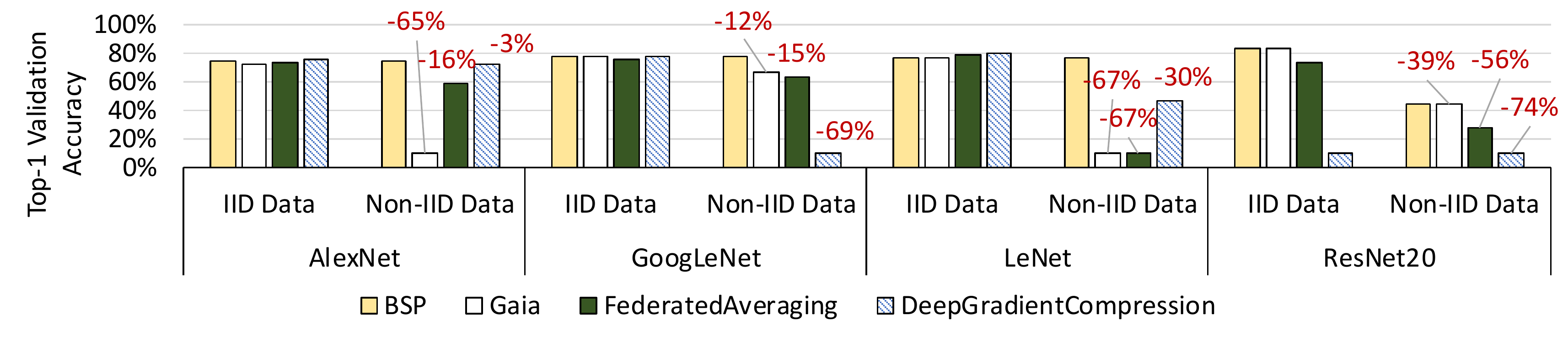}\vspace{-0.2in}
\caption{Top-1 validation accuracy for \appimage over the CIFAR-10 dataset. 
\pg{A} ``-x\%'' label \pg{above a bar} indicates the accuracy loss \pg{relative} to BSP in the IID setting.}
\label{fig:overview_cifar10} 
\end{figure*}

We first present the model quality with different decentralized learning algorithms \pg{in} the IID and Non-IID settings for \appimage \kv{using} the CIFAR-10 dataset. 
We use five partitions ($K\!\!=\!\! 5$) in this evaluation, \kv{and we discuss results with more partitions in Appendix~\ref{sec:more_partition}.}
As the CIFAR-10 dataset consists of ten object classes, each data partition has two object classes in the Non-IID setting. 
Figure~\ref{fig:overview_cifar10} shows the results with four popular DNNs (AlexNet, GoogLeNet, LeNet, and ResNet).
\pg{(Convergence curves for AlexNet and ResNet are shown in Appendix~\ref{appendix:convergence_curves}.)}
According to the hyper-parameter criteria in \xref{sec:setup}, we select $T_0 = 10\%$ for \gaia, $Iter_{Local} = 20$ for \fedavg, and $E_{warm} = 8 $ for \dgc.
We make two major observations.

\textbf{1) Non-IID data is a pervasive problem.} 
\emph{All} three decentralized learning algorithms lose significant model quality for \emph{all} four DNNs in the Non-IID setting. 
We see that while these algorithms retain the validation accuracy of BSP in the \textit{IID setting} with 15$\times$--20$\times$ communication savings (agreeing with the results from the original papers for these algorithms), they lose 3\% to 74\% validation accuracy in the \textit{Non-IID setting}.
Simply running these algorithms for more epochs would not help because the training/validation accuracy has already stopped improving.
Furthermore, the training completely \kv{diverges} in some cases, such as \dgc with GoogLeNet and ResNet20 (\dgc with ResNet20 also diverges in the IID setting). 
The pervasiveness of the problem is quite surprising, as we have a diverse set of decentralized learning algorithms and DNNs. 
This result shows that Non-IID data is a pervasive and challenging problem for decentralized learning, and this problem has been heavily understudied. 
\xref{subsec:cause} discusses \kv{potential causes} of this problem.

\textbf{2) Even BSP cannot completely solve this problem.} We see that even BSP, with its full communication \kv{at} every step, cannot retain model quality for some DNNs in the Non-IID setting. 
The validation accuracy of ResNet20 in the Non-IID setting is 39\% lower than that in the IID setting. 
This finding suggests that, for some DNNs, it \emph{may not be possible} to solve the Non-IID data challenge by increasing communication between data partitions. 
We find that this problem exists not only in ResNet20, but \kv{also in all other} DNNs we study with batch normalization (ResNet10, BN-LeNet~\cite{DBLP:conf/icml/IoffeS15} and Inception-v3~\cite{DBLP:conf/cvpr/SzegedyVISW16}).
We discuss this problem and potential solutions in \xref{sec:batch_norm}.

\textbf{The same trend in a larger dataset.} We conduct a similar study \pg{using} the ImageNet dataset~\cite{ILSVRC15} (1,000 image classes). 
We observe the same problems in the ImageNet dataset (e.g., an 8.1\% to 61.7\% accuracy loss on ResNet10), whose number of classes is two orders of magnitude more than the CIFAR-10 dataset. Appendix~\ref{sec:appendix_imagenet} discusses the experiment in detail.

\begin{figure}[t]
\centering
\includegraphics[width=0.47\textwidth]{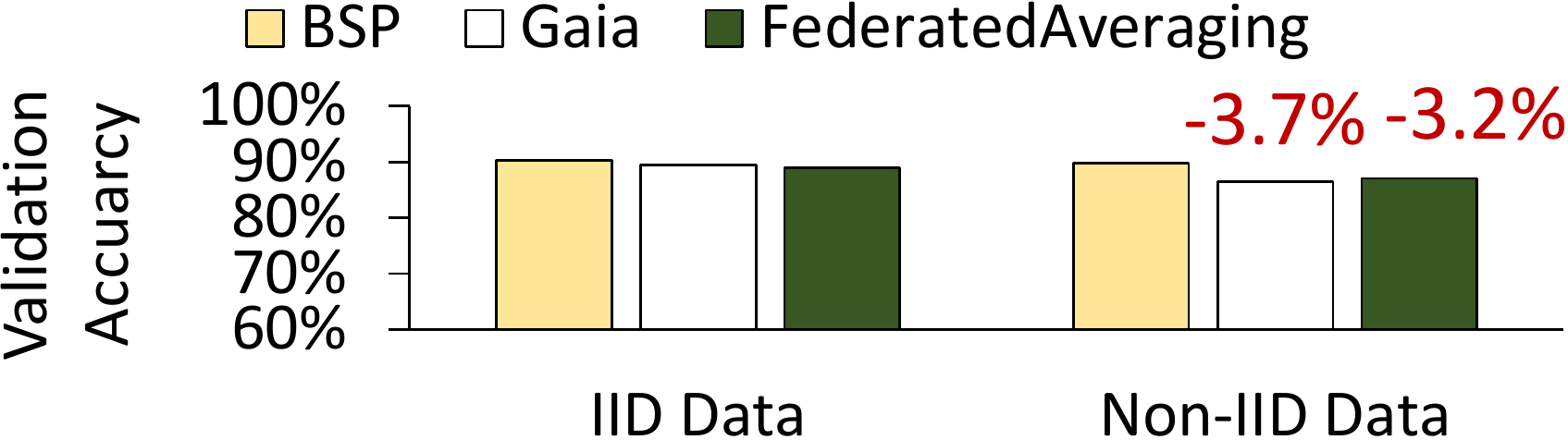}
\caption{\pg{GoogLeNet's} Top-1 validation accuracy for \appimage over the \FlickrMammal dataset, where 5\% data are randomly selected as the validation set. 
Non-IID Data is based on real-world data distribution among continents, and IID Data is the artificial setting in which \pg{images} are randomly assigned to partitions.
Each ``-x\%'' label indicates the accuracy loss \pg{relative} to BSP in the IID setting. 
Note: The y-axis starts at 60\% accuracy.} 
\label{fig:overview_result_geo_animal}
\end{figure}

\textbf{The same problem in real-world datasets.} We run similar experiments on our \FlickrMammal dataset. 
We use five partitions ($K\!\!=\!\! 5$) in this experiment, one for each continent, where each partition has as its local training data precisely the images from its corresponding continent.
Thus, we capture the real-world non-IID setting present in \FlickrMammal.
For comparison, we also consider an artificial IID setting, in which all the \FlickrMammal images are randomly distributed among the five partitions.
Figure~\ref{fig:overview_result_geo_animal} shows the results.
\pg{We use GoogLeNet in this experiment, and we select $T_0 = 10\%$ for \gaia and $Iter_{Local} = 20$ for \fedavg based on the criteria in \xref{sec:setup}.}
We observe the same problems for decentralized learning algorithms on this real-world dataset.
Specifically, \gaia and \fedavg are able to retain the model quality in the (artificial) IID setting, but they lose 3.7\% and 3.2\% accuracy in the (real-world) Non-IID setting, respectively.
\kv{The loss is smaller compared to Figure~\ref{fig:overview_cifar10} in part because most data labels still exist in all data partitions in the (real-world) Non-IID setting, which makes the problem easier than the 100\% non-IID setting.}
This loss arises even with modest hyper-parameter settings, and is expected to be larger with settings that more greatly reduce communication.
\kv{We also show that the loss increases to 5.2\% and 5.5\%, respectively, when \pg{\FlickrMammal} is partitioned at the subcontinent level (Appendix~\ref{sec:more_partition}).}
This is significant as the result suggests that skewed labels arising in real-world settings are a major problem for decentralized learning.

\subsection{Face Recognition}
\label{subsec:overview_face}

We further examine another popular ML application, \appface, to see if the Non-IID data problem is a challenge across different applications. 
We use two partitions in this evaluation. 
According to the hyper-parameter criteria in \xref{sec:setup}, we select $T_0\!\!=\!\! 20\%$ for \gaia and $Iter_{Local}\!\!=\!\! 50$ for \fedavg.
It is worth noting that the verification process of \appface is fundamentally different from \appimage, as \appface does \emph{not} use the classification layer (and thus the training labels) at all in the verification process. 
Instead, for each pair of verification images, the DNN uses the distance between the feature vectors of these images to determine \pg{whether the two images are of} the same person.

\begin{figure}[b]
\centering
\includegraphics[width=0.45\textwidth]{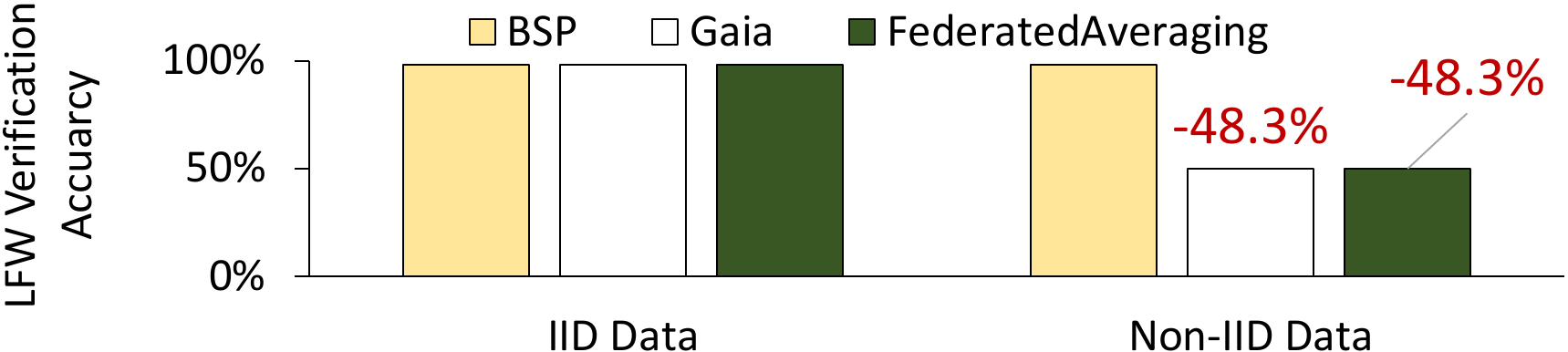}
\caption{LFW verification accuracy for \appface. Labels show the accuracy loss \pg{relative} to BSP in the IID setting.} 
\label{fig:overview_face}
\end{figure}

\textbf{The same problem in different applications.} Figure~\ref{fig:overview_face} shows the LFW verification accuracy. Again, the same problem happens: the decentralized learning algorithms work well in the IID setting, but they lose significant accuracy in the Non-IID setting. In fact, both \gaia and \fedavg cannot converge to a useful model in the Non-IID setting: their 50\% accuracy is no better than random guessing for the binary questions. \kv{This result is interesting because the labels of the validation dataset are completely different from the labels of the training dataset, but the validation accuracy is still severely impacted by the non-IID data label partitions in the training set.}

\subsection{The Problem of Decentralized Algorithms}
\label{subsec:cause}

The above results show that three diverse decentralized learning algorithms all \kv{experience} drastic accuracy \kv{losses} in the Non-IID setting.
We find two reasons for the accuracy loss.
First, for algorithms such as \gaia that save communication by allowing small model differences in each \kv{partition} $P_k$, the Non-IID setting results in \emph{completely different models among $P_k$}.
The small differences give local models room for specializing \pg{to} local data. 
Second, for algorithms that save communication by synchronizing sparsely (e.g., \fedavg and \dgc), each $P_k$ generates more diverged gradients in the Non-IID setting, which is not surprising as each $P_k$ sees vastly different training data. 
When they are finally synchronized, they may have diverged so much from the global model that they \pg{push} the global model the wrong direction. See Appendix~\ref{sec:decentral_cause} for \pg{further details}.

\subsection{Algorithm Hyper-Parameters}
\label{subsec:decentral_parameter}

We also study the sensitivity of the Non-IID problem to hyper-parameter choice among decentralized learning algorithms. 
We find that even relatively conservative hyper-parameter settings, which incur high communication costs, still suffer major accuracy loss in the Non-IID setting. 
In the IID setting, on the other hand, the \textit{same} hyper-parameter achieves similar high accuracy as BSP. 
In other words, the the Non-IID problem is not specific to particular hyper-parameter choices. Appendix~\ref{sec:decentral_parameter} shows \kv{supporting} results.

\section{Batch Normalization: Problem and Solution}
\label{sec:batch_norm}


\subsection{Batch Normalization in the Non-IID Setting}
\label{subsec:batch_norm_problem}


\textbf{How BatchNorm works.} 
Batch normalization (BatchNorm)~\cite{DBLP:conf/icml/IoffeS15} is one of the most popular mechanisms in deep learning (20,000+ citations as of August 2020). 
BatchNorm aims to stabilize a DNN by normalizing the input distribution to zero mean and unit variance. 
Because the \textit{global} mean and variance are unattainable with stochastic training, BatchNorm uses \emph{minibatch mean and variance} as an estimate of the global mean and variance. 
Specifically, for each minibatch $\mathcal{B}$, BatchNorm calculates the minibatch mean $\mu_{\mathcal{B}}$ and variance $\sigma_{\mathcal{B}}$, and then uses $\mu_{\mathcal{B}}$ and $\sigma_{\mathcal{B}}$ to normalize each input in $\mathcal{B}$. 
BatchNorm enables faster and more stable training because it enables larger learning rates~\cite{DBLP:conf/nips/BjorckGSW18, DBLP:conf/nips/SanturkarTIM18}.

\textbf{BatchNorm and the Non-IID setting.} 
While BatchNorm is effective in practice, its dependence on minibatch mean and variance ($\mu_{\mathcal{B}}$ and $\sigma_{\mathcal{B}}$) is known to be problematic in certain settings. 
This is because BatchNorm uses $\mu_{\mathcal{B}}$ and $\sigma_{\mathcal{B}}$ for training, but it typically uses an estimated global mean and variance ($\mu$ and $\sigma$) for validation. 
If there is a major mismatch between these means and variances, the validation accuracy is going to be low.
This can happen if the minibatch size is small or the sampling of minibatches is not IID~\cite{DBLP:conf/nips/Ioffe17}. 
The Non-IID setting in our study exacerbates this problem because each data partition $P_k$ sees very different training samples. 
Hence, the $\mu_{\mathcal{B}}$ and $\sigma_{\mathcal{B}}$ in each \pg{partition can vary significantly across the partitions}, and the synchronized global model may not work for \emph{any} set of data.
Worse still, we cannot simply increase the minibatch size or do better minibatch sampling to solve this problem, because in the Non-IID setting the underlying dataset in each $P_k$ does not represent the global dataset.

\begin{figure}[h]
\centering
\includegraphics[width=0.4\textwidth]{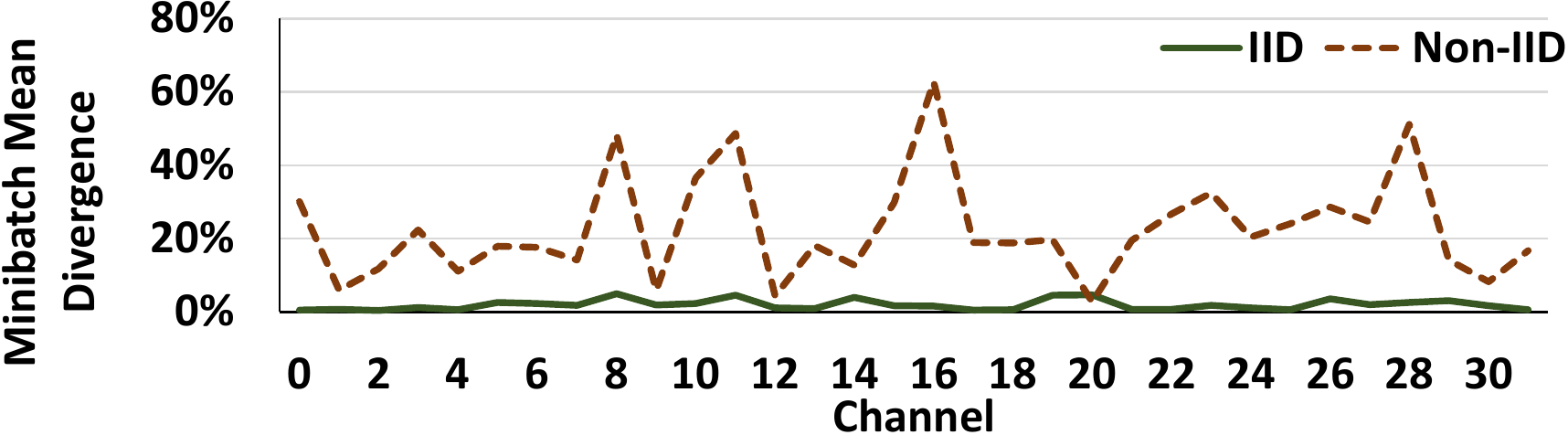}\vspace{-0.1in}
\caption{Minibatch mean divergence for the first layer of BN-LeNet over CIFAR-10 using two $P_k$.} 
\label{fig:batch_mean_divergence}
\vspace{-0.1in}
\end{figure}

We validate if there is indeed major divergence in $\mu_{\mathcal{B}}$ and $\sigma_{\mathcal{B}}$ among different $P_k$ in the Non-IID setting. 
We calculate the divergence of $\mu_{\mathcal{B}}$ as the difference between $\mu_{\mathcal{B}}$ in different $P_k$ over the average $\mu_{\mathcal{B}}$ (i.e., it is $\frac{||\mu_{\mathcal{B}, P_0} - \mu_{\mathcal{B}, P_1}||}{||AVG(\mu_{\mathcal{B}, P_0},\ \mu_{\mathcal{B}, P_1})||}$ for two partitions $P_0$ and $P_1$). 
We use the average $\mu_{\mathcal{B}}$ over every 100 minibatches in each $P_k$ so that we get better estimation. 
Figure~\ref{fig:batch_mean_divergence} depicts the divergence of $\mu_{\mathcal{B}}$ for each channel of the first layer of BN-LeNet, which is constructed by inserting BatchNorm to LeNet after each convolutional layer. 
As we see, the divergence of $\mu_{\mathcal{B}}$ is significantly larger in the Non-IID setting (between 6\% to \pg{61\%}) than in the IID setting (between 1\% to 5\%). 
We also observe the same trend in minibatch variances $\sigma_{\mathcal{B}}$ (not shown).
As this problem has nothing to do with the amount of communication among $P_k$, it explains why even BSP cannot retain model accuracy for BatchNorm in the Non-IID setting.

\subsection{Alternatives to Batch Normalization}
\label{subsec:batch_norm_solution}

As the problem of BatchNorm in the Non-IID setting is due to its dependence on minibatches, the natural solution is to replace BatchNorm with alternative normalization mechanisms that are \emph{not} dependent on minibatches. 
Unfortunately, most existing alternative normalization mechanisms (Weight Normalization~\cite{DBLP:conf/nips/SalimansK16}, Layer Normalization~\cite{DBLP:journals/corr/BaKH16}, Batch Renormalization~\cite{DBLP:conf/nips/Ioffe17}) have their own drawbacks (see Appendix~\ref{appendix:batchnorm}). 
Here, we discuss a particular mechanism that may be used instead. 

\textbf{Group Normalization.} 
Group Normalization (GroupNorm)~\cite{DBLP:conf/eccv/WuH18} is an alternative normalization mechanism that aims to overcome the shortcomings of BatchNorm and Layer Normalization (LayerNorm). 
GroupNorm divides adjacent channels into groups of a prespecified size $\mathcal{G}_{size}$, and computes the per-group mean and variance for \emph{each input sample}. 
Hence, GroupNorm does not depend on minibatches for normalization (the shortcoming of BatchNorm), and GroupNorm does not assume all channels make equal contributions (the shortcoming of LayerNorm). 

\begin{figure}[h]
\centering
\includegraphics[width=0.48\textwidth]{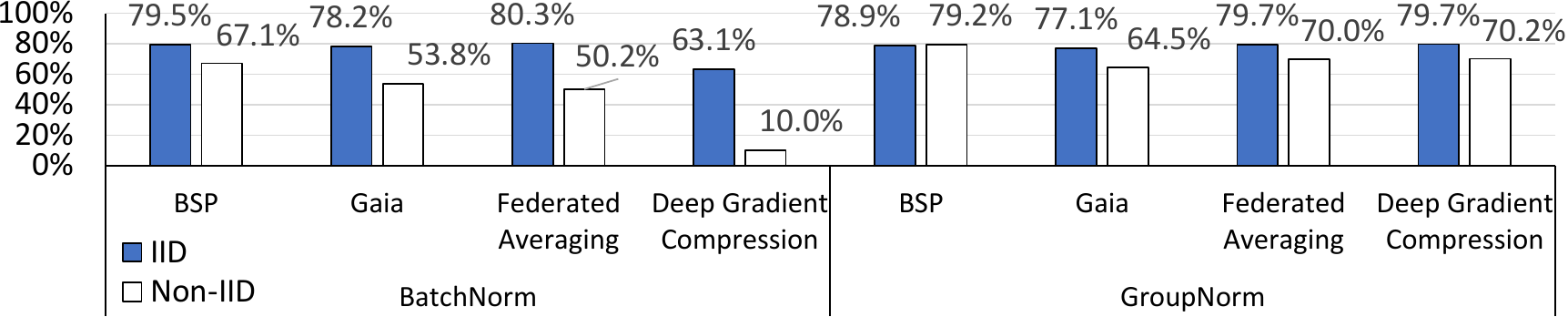}
\caption{Top-1 validation accuracy with BatchNorm and GroupNorm for BN-LeNet over CIFAR-10 with 5 partitions.} 
\label{fig:group_norm_results}
\end{figure}

We evaluate GroupNorm with BN-LeNet over CIFAR-10. 
We carefully select $\mathcal{G}_{size} = 2$, which works best with this DNN. 
Figure~\ref{fig:group_norm_results} shows the Top-1 validation accuracy with GroupNorm and BatchNorm across decentralized learning algorithms. 
We make two major observations.

First, GroupNorm successfully recovers the accuracy loss of BatchNorm with BSP in the Non-IID setting. 
As the figure shows, GroupNorm with BSP achieves 79.2\% validation accuracy in the Non-IID setting, which is as good as the accuracy in the IID setting. 
This shows GroupNorm can be used as an alternative to BatchNorm to overcome the Non-IID data challenge for BSP. 
Second, GroupNorm dramatically helps the decentralized learning algorithms in the Non-IID setting as well.
With GroupNorm, there is 14.4\%, 8.9\% and 8.7\% accuracy loss for \gaia, \fedavg and \dgc, respectively. 
While the accuracy losses are still significant, they are better than their BatchNorm counterparts by an additive 10.7\%, 19.8\% and 60.2\%, respectively. 

\textbf{Discussion.} 
While our study shows that GroupNorm can be a good alternative to BatchNorm in the Non-IID setting,
it is worth noting that BatchNorm is widely adopted in many DNNs. Hence, more study \kv{is needed} to see if GroupNorm can replace BatchNorm for different applications and DNN models. 
As for other tasks such as recurrent (e.g., LSTM~\cite{DBLP:journals/neco/HochreiterS97}) and generative (e.g., GAN~\cite{DBLP:conf/nips/GoodfellowPMXWOCB14}) models, other normalization techniques such as LayerNorm
can be good options because \emph{(i)} they are shown to be effective in these tasks and \emph{(ii)} they are not dependent on minibatches.

\section{Degree of \kv{Data Skew}}
\label{sec:skewness}

\begin{figure}[h]
\vspace{0.1in}
\centering  
\includegraphics[width=0.48\textwidth]{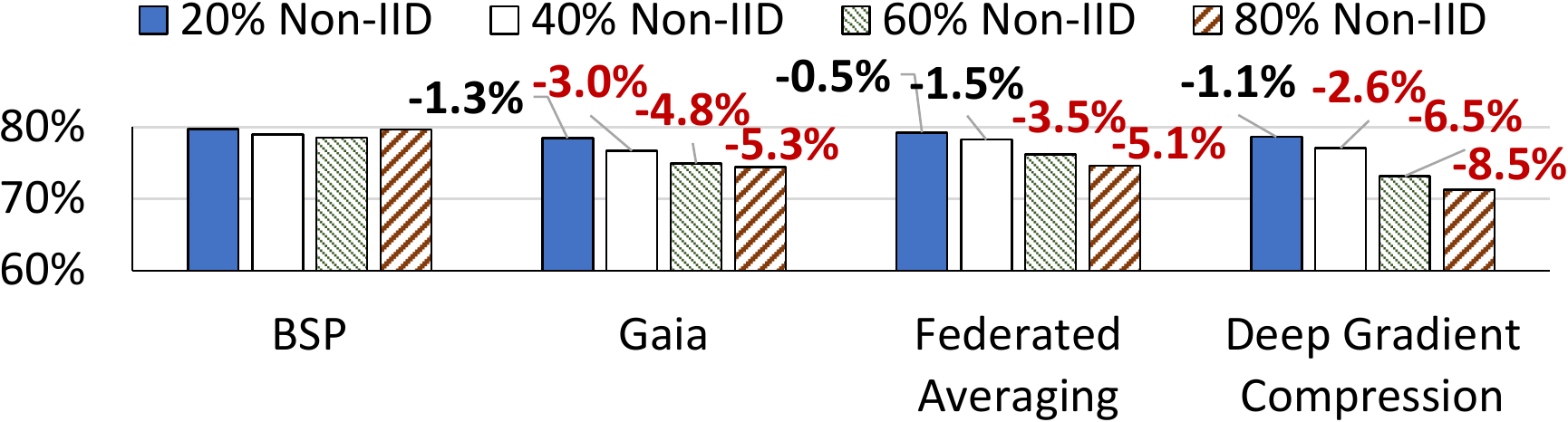}
\caption{Top-1 validation accuracy for GN-LeNet over CIFAR-10, varying the degree of skew. 
\pg{Each ``-x\%'' label indicates} the accuracy loss \pg{relative} to BSP in the IID setting. Note: \pg{The} y-axis starts at 60\% accuracy.}
  \label{fig:noniid_degree}
\end{figure}

In \xref{sec:overview}--\xref{sec:batch_norm}, we studied a strict case of skewed label partitions, where each label only exists in a \kv{single} data partition, \emph{exclusively}
\pg{(the one exception being our experiments with \FlickrMammal)}.
While this case may be a reasonable approximation for some applications (e.g., for \appface, a person's face image may exist only in one data partition), it could be an extreme case for other applications (e.g., \appimage, as \xref{subsec:dataset} shows).
Here, we study how the problem changes with the degree of skew by controlling the fraction of \pg{the dataset that is} non-IID (\kv{i.e., partitioned using labels}, \xref{sec:setup}). 
Figure~\ref{fig:noniid_degree} \kv{shows} the CIFAR-10 validation accuracy of GN-LeNet (our name for BN-LeNet with GroupNorm replacing BatchNorm) in the 20\%, 40\%, 60\% and 80\% non-IID setting.
We make two observations.

\textbf{1) Partial non-IID data is also problematic.} 
We see that for all three decentralized learning algorithms, partial non-IID data \kv{can} still cause major accuracy loss. 
Even with a small degree of non-IID data such as 40\%, we still see 1.5\%--3.0\% accuracy loss.
Thus, the problem of non-IID data does not occur only with exclusive label partitioning, and the problem exists in \kv{the} vast majority of practical settings. 

\textbf{2) Degree of skew often determines the difficulty level of the problem.} 
The model accuracy gets worse with higher degrees of skew, and the accuracy gap between 80\% and 20\% non-IID data can be as large as 7.4\% (\dgc). 
In general, we see that the problem \pg{becomes} more difficult with higher \kv{degree of} skew.

\section{Our Approach: {\nameSec}}
\label{sec:solution}

To address the problem of skewed label partitions, we introduce {\name}, a general approach that enables communication-efficient decentralized learning over \emph{arbitrarily} skewed label partitions. 

\subsection{Overview of \name}
\label{subsec:solution_overview}
  
\begin{figure}[t]
\centering
\includegraphics[width=0.4\textwidth]{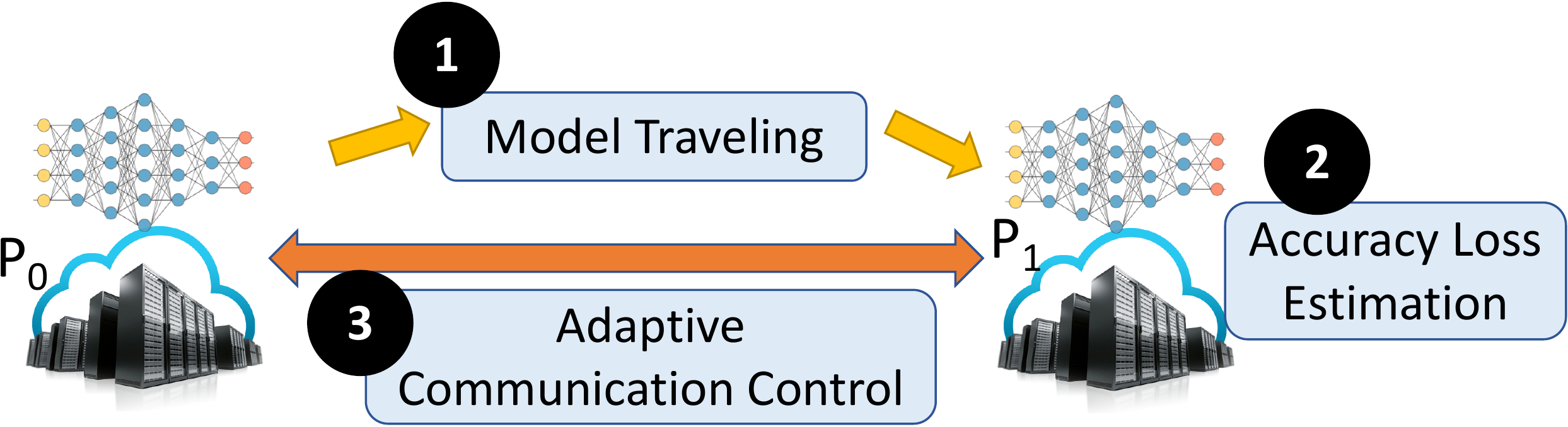}
\caption{Overview of {\name}}
\label{fig:solution_overview}
\end{figure}

We design {\name} as a general module that can be seamlessly integrated with different decentralized learning algorithms, ML training frameworks, and ML applications. Figure~\ref{fig:solution_overview} \kv{provides an overview of} the {\name} design.

\begin{enumerate}[topsep=0pt,itemsep=0ex,partopsep=0ex,parsep=1ex,leftmargin=3ex]

\item \textbf{\kv{Estimating} the degree of skew.} As \xref{sec:skewness} shows, knowing the degree of skew is very useful to determine an appropriate solution.
  To learn this information, {\name} periodically moves the ML model from one data partition ($P_k$) to another during training (\emph{model traveling}, \ding{182} in Figure~\ref{fig:solution_overview}).
  {\name} then evaluates how well a model performs on a remote data partition by evaluating the model accuracy with a subset of training data on the remote node. As we already know the training accuracy of this model in its original data partition, we can infer the \emph{accuracy loss} in the remote data partition (\ding{183}).

\item \textbf{Adaptive communication control (\ding{184}).} Based on the \kv{estimated} accuracy loss, {\name} controls the amount of communication among data partitions to retain model quality. {\name} controls the amount of communication by automatically tuning the hyper-parameters of the decentralized learning algorithm (\xref{subsec:decentral_parameter}). 
  This tuning process essentially \kv{solves} an optimization problem that aims to minimize communication among data partitions while keeping accuracy loss within a reasonable threshold (\pg{further details below}).

\end{enumerate}

In essence, {\name} handles non-IID data partitions by controlling communication based on accuracy loss.
{\name} is agnostic to the source of the loss, which may be \kv{due to} skewed label partitions or other forms of non-IID data (Appendix~\ref{appendix:discussion}).
As long as increasing communication improves accuracy \kv{for the data skew}, {\name} should be effective \kv{in retaining model quality while minimizing communication.}

\subsection{Mechanism Details}
\label{subsec:comm_control}

We discuss the mechanisms of {\name} in detail.

\noindent \textbf{Accuracy Loss.} The accuracy loss between data partitions represents the degree of model divergence. As \xref{subsec:cause} discusses, ML models in different data partitions tend to specialize for their training data, especially when we use decentralized learning algorithms to reduce communication. 

We study accuracy loss under \gaia, for hyper-parameter choices $T_0\!\!=\!\! 2\%, 5\%, 10\%, 20\%$, in the IID and non-IID settings.
We find that accuracy loss changes drastically from the IID setting (0.4\% on average) to the Non-IID setting (39.6\% on average), and that lower $T_0$ results in smaller accuracy loss in the non-IID setting.
See Appendix~\ref{appendix:accuracy_loss} for \pg{further details}.
Accordingly, we can use accuracy loss (i) to estimate how much the models diverge from each other (reflecting training data differences); and (ii) to serve as an objective function for communication control.
The computation overhead to evaluate accuracy loss is quite small because we run inference with only a small fraction of training data, and we only do so once in a while (we empirically find that once every 500 mini-batches is frequent enough).


\noindent \textbf{Communication Control.} The goal of communication control is to retain model quality while minimizing communication among data partitions. We achieve this by solving an optimization problem, which aims to minimize communication while keeping the \emph{accuracy loss} \kv{below} a small threshold $\sigma_{AL}$ so that we can control model divergence caused by non-IID data partitions. We solve this optimization problem periodically after we estimate the accuracy loss with model \pg{traveling}. Specifically, our target function is:

\ignore{
Specifically, given a set of hyper-parameters $\theta_t$ for each iteration (or minibatch) $t$, the optimization problem for {\name} is to minimize the total amount of communication for a data partition $P_k$:

\ignore{
\begin{small}
  \begin{equation}
    \text{Comm} = \sum_{mt=0}^{\ceil*{\frac{T(\theta)}{MTP}}} \sum_{t=mt \cdot MTP}^{(mt + 1) \cdot MTP} C(\theta_t) + \sum_{mt=0}^{\ceil*{\frac{T(\theta)}{MTP}}} CM
  \end{equation}
\end{small}
}

\begin{small}
  \begin{equation}
    \label{eq:opt}
    \argmin_{\theta, MTP} \left( \sum_{mt=0}^{\ceil*{\frac{T(\theta)}{MTP}}} \sum_{t=mt \cdot MTP}^{(mt + 1) \cdot MTP} C(\theta_t) + \sum_{mt=0}^{\ceil*{\frac{T(\theta)}{MTP}}} CM \right)
  \end{equation}
\end{small}

where $T(\theta)$ is the total number of iterations to achieve the target model accuracy given all hyper-parameters $\theta$ throughout the training, $C(\theta_t)$ is the amount of communication given $\theta_t$, $MTP$ is the period size (in iterations) for model traveling, and $CM$ is the communication cost for the ML model (for model traveling).
}



\begin{small}
  \vspace{-15pt}
  \begin{equation}
  \label{eq:tune}
    \argmin_{\theta} \left( \lambda_{AL} \left( \texttt{max}(0, AL (\theta) - \sigma_{AL}) \right) + \lambda_{C} \frac{C(\theta)}{CM} \right)
  \end{equation}
  \vspace{-15pt}
\end{small}

where $AL (\theta)$ is the accuracy loss based on the previously selected hyper-parameter $\theta$ (we memoize the most recent value for each $\theta$ that has been explored), $C(\theta)$ is the amount of communication given $\theta$, $CM$ is the communication cost for the whole ML model, and $\lambda_{AL}$, $\lambda_{C}$ are given parameters to determine the weights of accuracy loss and communication, respectively. We can employ various algorithms with Equation~\ref{eq:tune} to select $\theta$, such as hill climbing, stochastic hill climbing~\cite{russell2020artificial}, and simulated annealing~\cite{van1987simulated}. 


\subsection{Evaluation Results}
\label{subsec:solution_results}

We implement and evaluate {\name} in a GPU parameter server system~\cite{DBLP:conf/eurosys/CuiZGGX16} based on Caffe~\cite{DBLP:journals/corr/JiaSDKLGGD14}. We evaluate several aforementioned auto-tuning algorithms and we find that hill climbing provides the best results.
As our primary goal is to minimize accuracy loss, we set $\lambda_{AL}=50$ and $\lambda_{AC}=1$.
We \kv{set} $\sigma_{AL}=5\%$ to tolerate an acceptable accuracy variation during training, which does not reduce the final validation accuracy.

\begin{figure}[t]
  \centering
  \begin{subfigure}[t]{0.48\linewidth}
    \centering
    \includegraphics[width=1.0\textwidth]{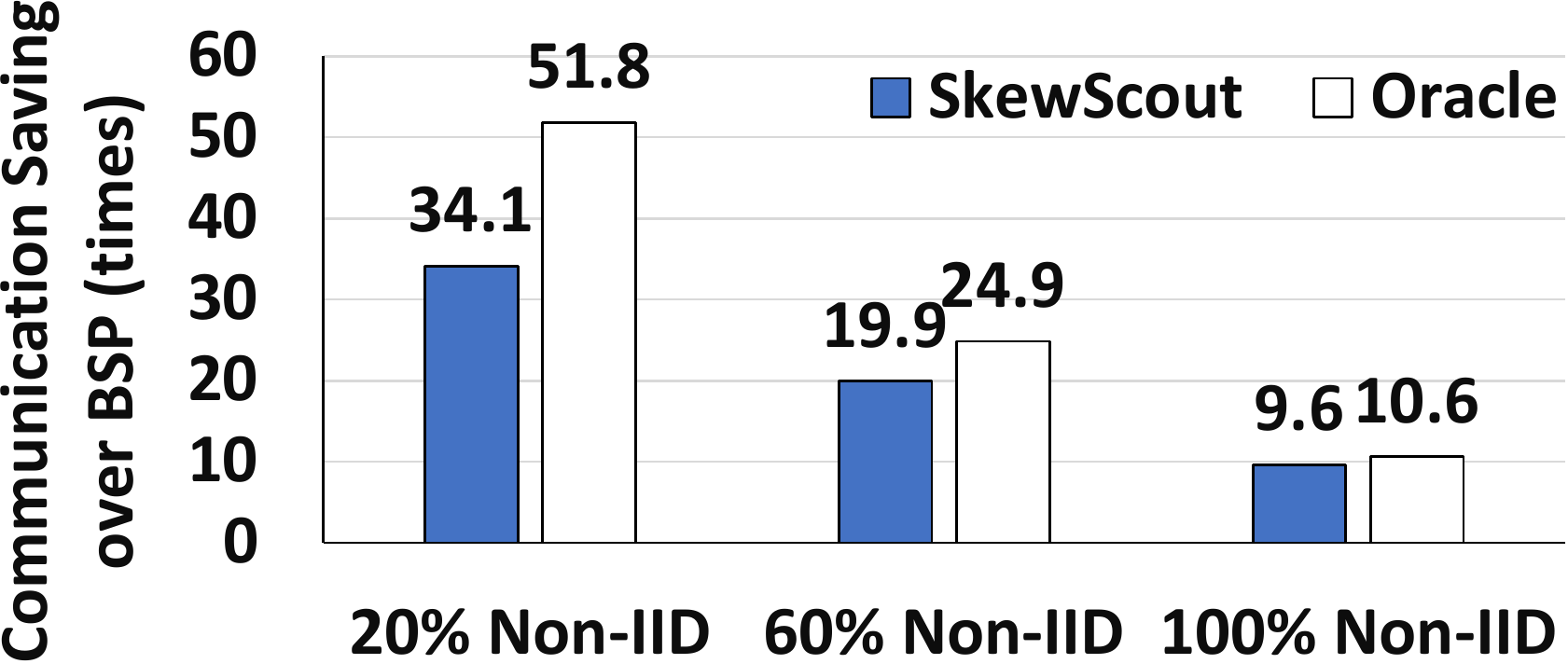}
    \caption{AlexNet}
  \end{subfigure}
   \centering 
  \begin{subfigure}[t]{0.48\linewidth}
    \centering
    \includegraphics[width=1.0\textwidth]{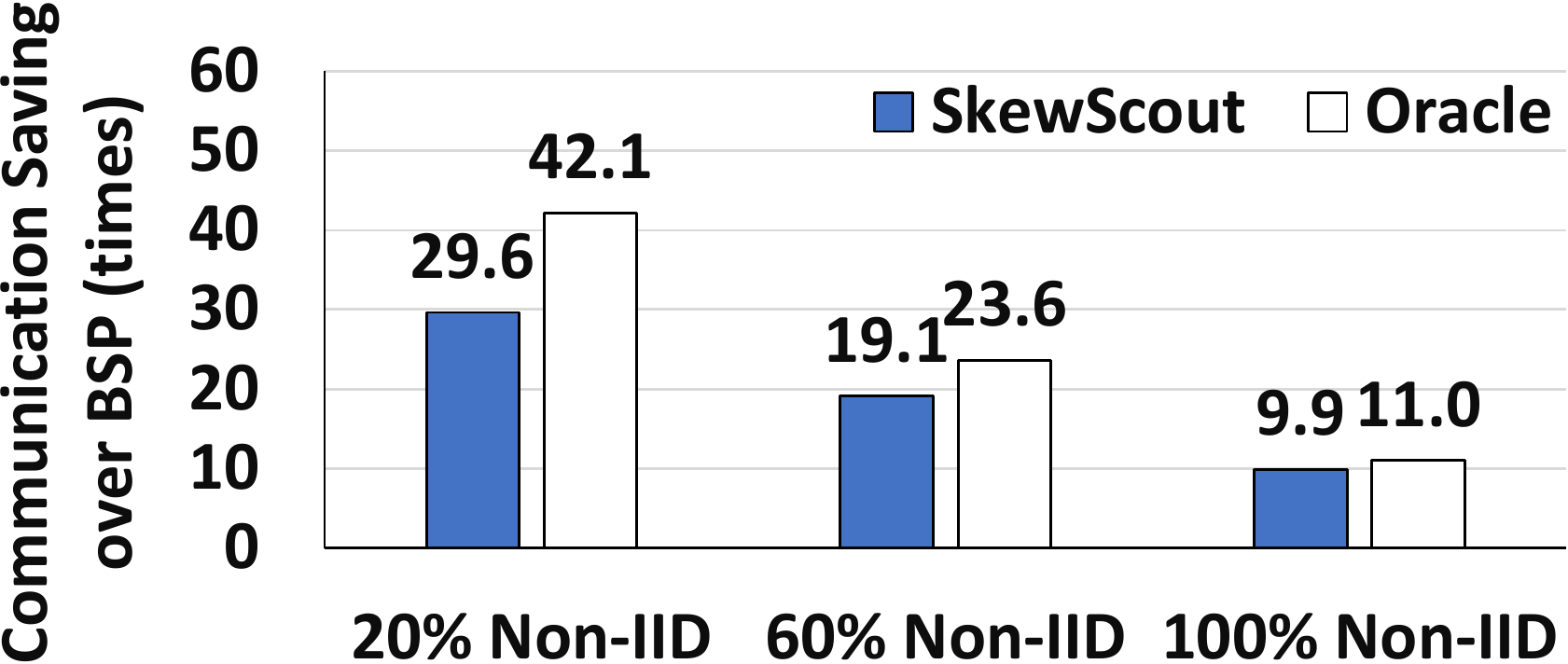}
    \caption{GoogLeNet}
  \end{subfigure}
  \vspace{-0.05in}
  \caption{Communication savings over BSP with {\name} and \sys{Oracle} for training \kv{with} CIFAR-10. All results achieve the same accuracy as BSP in the IID setting.}
  \label{fig:solution_result}
\end{figure}

We compare {\name} with two other baselines: (1) \sys{BSP}: the most communication-heavy approach that retains model \kv{accuracy} in all Non-IID settings; and (2) \sys{Oracle}: the ideal, yet unrealistic, approach that selects the most communication-efficient $\theta$ that retains model \kv{accuracy}, by \emph{running all possible $\theta$} in each setting prior to measured execution. Figure~\ref{fig:solution_result} shows the communication savings over \sys{BSP} for both {\name} and \sys{Oracle} when training with \gaia. Note that all results achieve \emph{the same} validation accuracy as \sys{BSP}. We make two observations.

First, {\name} is much more effective than \sys{BSP} in handling Non-IID settings. Overall, {\name} achieves 9.6--34.1$\times$ communication savings over \sys{BSP} in various Non-IID settings without sacrificing model accuracy. As expected, {\name} saves more communication with less skewed data because {\name} can safely loosen communication. 

Second, {\name} is not far from the ideal \sys{Oracle} baseline. Overall, {\name} requires only 1.1--1.5$\times$ more communication than \sys{Oracle} to achieve the same model accuracy. {\name} cannot match the communication savings of \sys{Oracle} because: (i) {\name} does model traveling periodically, which leads to some overhead; and (ii) for some $\theta$, high accuracy loss at the beginning can still \kv{lead to} a high accuracy model, which {\name} cannot foresee. As \sys{Oracle} requires \textit{many} runs in practice, we conclude that {\name} is an effective, \kv{realistic} one-pass solution for decentralized learning over non-IID data partitions. 

\section{Related Work}
\label{sec:related}

To our knowledge, this is the first study to show that skewed label partitions across devices/locations is a fundamental and pervasive problem for decentralized learning. 
Our study investigates various aspects of this problem, such as a real-world dataset, decentralized learning algorithms, batch normalization, and data skew, as well as presenting our \name{} approach.
Here, we discuss related work.

\textbf{Large-scale systems for centralized learning.} 
There are many large-scale ML systems that aim to enable efficient ML training over \emph{centralized} datasets using communication-efficient designs, such as relaxing synchronization requirements~\cite{DBLP:conf/nips/RechtRWN11, HoCCLKGGGX13, DBLP:journals/corr/GoyalDGNWKTJH17} or sending fewer updates to parameter servers~\cite{DBLP:conf/osdi/LiAPSAJLSS14, DBLP:conf/nips/LiASY14}. 
These works assume the training data are centralized so they can be easily partitioned among the machines performing the training in an IID manner (e.g., by random shuffling). 
Hence, they are neither designed \kv{for} nor validated on non-IID data partitions.

\ignore{
\textbf{Communication-efficient training for specific algorithms.} 
A large body of prior work proposes ML training algorithms to reduce the dependency on intensive parameter updates to enable more efficient parallel training (e.g.,~\cite{DBLP:conf/nips/JaggiSTTKHJ14, DBLP:journals/jmlr/ZhangDW13, DBLP:conf/nips/ZinkevichWSL10, DBLP:conf/icml/TakacBRS13, DBLP:conf/uai/NeiswangerWX14, DBLP:conf/icml/ShamirS014, DBLP:conf/icml/ZhangL15a, DBLP:journals/jmlr/Shalev-Shwartz013}). 
These works reduce communication overhead by proposing algorithm-specific approaches, such as solving a dual problem~\cite{DBLP:journals/jmlr/Shalev-Shwartz013, DBLP:conf/icml/TakacBRS13, DBLP:conf/nips/JaggiSTTKHJ14} or employing a different optimization algorithm~\cite{DBLP:conf/icml/RakhlinSS12, DBLP:conf/icml/ZhangL15a}. 
The main drawback of these approaches is that they are not general and their applicability depends on the ML application. 
Besides, these works also assume centralized, IID data partitions. 
Thus, their effectiveness over decentralized, non-IID data partitions needs much more study.
}

\textbf{Decentralized learning.} 
Recent prior work proposes communication-efficient algorithms (e.g.,~\cite{DBLP:conf/nsdi/HsiehHVKGGM17, DBLP:conf/aistats/McMahanMRHA17, DBLP:conf/ccs/ShokriS15, DBLP:conf/ICLR/LinHMWD18, DBLP:conf/icml/TangLYZL18}) for ML training over \emph{decentralized} datasets. 
However, as our study shows, these decentralized learning algorithms lose significant model \kv{accuracy} in the Non-IID setting (\xref{sec:overview}). 
Some recent work studies the problem of non-IID data partitions.
For example, instead of training a global model to fit non-IID data partitions, federated multi-task learning~\cite{DBLP:conf/nips/SmithCST17} trains local models for each data partition while leveraging other data partitions to improve model accuracy. 
However, this approach sidesteps the problem for global models, which are essential when a local model is unavailable (e.g., a brand new partition without training data) or ineffective (e.g., a partition with too few training examples for a class).
Several recent works show significant accuracy loss for \fedavg over non-IID data, and some propose algorithms to improve \fedavg over non-IID data~\cite{DBLP:journals/corr/abs-1806-00582, DBLP:journals/corr/abs-1907-02189, DBLP:journals/corr/abs-1910-07796, DBLP:journals/corr/abs-1910-06378, DBLP:journals/corr/abs-1912-12844, DBLP:conf/mlsys/LiSZSTS20, DBLP:conf/iclr/WangYSPK20, DBLP:conf/aistats/0001MR20}.
While the result of these works aligns with our observations, our study \emph{(i)} broadens the problem scope to a variety of decentralized learning algorithms, ML applications, DNN models, and datasets, \emph{(ii)} explores the problem of batch normalization and possible solutions, and \emph{(iii)} designs and evaluates \name, which can also complement \kv{the aforementioned} algorithms by controlling their hyper-parameters over arbitrarily skewed data partitions. 

\textbf{Non-IID dataset.}
Recent work offers non-IID datasets to facilitate the study of federated learning. 
For example, LEAF~\cite{DBLP:journals/corr/abs-1812-01097} provides datasets that are partitioned in various ways.
Luo et al.~release 900 images collected from cameras in different locations, and they show severe skewed label distribution across cameras~\cite{DBLP:journals/corr/abs-1910-11089}.
Our study on geo-tagged mammals on Flickr shows the same problem at a much larger scale, and our dataset broadens the scope to include geo-distributed learning.

\section{Conclusion}

As most timely and relevant ML data 
\pg{are} generated at different \kv{physical locations}, and often infeasible/impractical to collect centrally, decentralized learning provides an important path for ML applications to leverage \kv{such} data. 
However, decentralized data \kv{is} often generated at different contexts, which leads to a heavily understudied problem: \emph{non-IID training data partitions}. 
We conduct a detailed empirical study of this problem for skewed label partitions, revealing three key findings. 
First, we show that training over skewed label partitions is a fundamental and pervasive problem for decentralized learning, as all decentralized learning algorithms in our study suffer major accuracy loss. 
Second, we find that DNNs with batch normalization are particularly vulnerable in the Non-IID setting, with even the most communication-heavy approach being unable to retain model quality. 
We further discuss the cause and a potential solution to this problem. 
Third, we show that the difficulty level of this problem varies greatly with the degree of skew. 
Based on these findings, we present \name, a general approach to
minimizing communication while retaining model quality even for non-IID data.
We hope that the findings and insights in this paper, as well as our open source code and dataset, will spur further research into the fundamental and important problem of non-IID data in decentralized learning.
\section*{Acknowledgments}

We thank H. Brendan McMahan, Gregory R. Ganger, and the anonymous ICML reviewers for their valuable and constructive suggestions. 
Special thanks to Chenghao Zhang for his help in setting up the face recognition experiments. 
We also thank the members and companies of the PDL Consortium (Alibaba, Amazon, Datrium, Facebook, Google, Hewlett-Packard Enterprise, Hitachi, IBM, Intel, Microsoft, NetApp, Oracle, Salesforce, Samsung, Seagate, and Two Sigma) for their interest, insights, feedback, and support.
We acknowledge the support of the \kv{SAFARI Research Group's} industrial partners: Google, Huawei, Intel, Microsoft, and VMware. 
This work is supported in part by NSF.
Most of the work was done when Kevin Hsieh was at Carnegie Mellon University.

  \bibliography{ref}
  \bibliographystyle{icml2020}

\appendix
\clearpage


\onecolumn

\section*{Appendix}

\section{Details of Decentralized Learning Algorithms}
\label{appendix:decentral_algo}

This section presents the pseudocode for \gaia, \fedavg, and \dgc.

\begin{algorithm}[h]
\caption{\gaia~\cite{DBLP:conf/nsdi/HsiehHVKGGM17} on node $k$ for vanilla momentum SGD}
\label{algo:gaia}
\begin{algorithmic}[1]
\Require initial weights $w_0 = \{w_0[0],..., w_0[M]\}$
\Require $K$ data partitions (or data centers); initial significance threshold $T_0$
\Require local minibatch size $B$; momentum $m$; learning rate $\eta$; local dataset $\mathcal{X}_k$
\State $u_0^k \gets 0$; $v_0^k \gets 0$
\State $w_0^k \gets w_0$
\For{$t = 0, 1, 2,...$}
\State $b \gets$ (sample $B$ data samples from $\mathcal{X}_k$)
\State $u_{t+1}^k \gets m \cdot u_t^k - \eta \cdot \bigtriangledown f(w_t^k, b)$
\State $w_{t+1}^k \gets w_{t}^k + u_{t+1}^k$
\State $v_{t+1}^k \gets v_{t}^k + u_{t+1}^k$ 
\Comment{Accumulate weight updates}
\For{$j = 0, 1,...M$}
\State $S \gets ||\frac{v_{t+1}^k}{w_{t+1}^k}|| > T_t$
\Comment{Check if accumulated updates are significant}
\State $\widetilde{v}_{t+1}^k[j] \gets v_{t+1}^k[j] \odot S$
\Comment{Share significant updates with other $P_k$}
\State $v_{t+1}^k[j] \gets v_{t+1}^k[j] \odot \neg S$
\Comment{Clear significant updates locally}
\EndFor
\For{$i = 0, 1,...K; i \neq k$}
\State $w_{t+1}^k \gets w_{t+1}^k + \widetilde{v}_{t+1}^i$
\Comment{Apply significant updates from other $P_k$}
\EndFor
\State $T_{t+1} \gets \texttt{update\_threshold}(T_t)$
\Comment{Decrease threshold whenever the learning rate decreases}
\EndFor
\end{algorithmic}
\end{algorithm}


\begin{algorithm}[h]
\caption{\fedavg~\cite{DBLP:conf/aistats/McMahanMRHA17} on node $k$ for vanilla momentum SGD}
\label{algo:fedavg}
\begin{algorithmic}[1]
\Require initial weights $w_0$; $K$ data partitions (or clients)
\Require local minibatch size $B$; local iteration number $Iter_{Local}$
\Require momentum $m$; learning rate $\eta$; local dataset $\mathcal{X}_k$
\State $u^k \gets 0$
\For{$t = 0, 1, 2,...$}
\State $w_t^k \gets w_t$
\Comment{Get the latest weights from the server}
\For{$i = 0,...Iter_{Local}$}
\State $b \gets$ (sample $B$ data samples from $\mathcal{X}_k$)
\State $u^k \gets m \cdot u^k - \eta \cdot \bigtriangledown f(w_t^k, b)$
\State $w_t^k \gets w_{t}^k + u^k$
\EndFor
\State $\texttt{all\_reduce:}\: w_{t+1} \gets \sum^{K}_{k=1} \frac{1}{K} w_t^k$
\Comment{Average weights from all partitions}
\EndFor
\end{algorithmic}
\end{algorithm}
\kv{In order make our experiments deterministic and simpler, we use all data partitions (or clients) in every epoch for \fedavg.}    
\clearpage

\begin{algorithm}[h]
\caption{\dgc~\cite{DBLP:conf/ICLR/LinHMWD18} on node $k$ for vanilla momentum SGD}
\label{algo:dgc}
\begin{algorithmic}[1]
\Require initial weights $w_0 = \{w_0[0],..., w_0[M]\}$
\Require $K$ data partitions (or data centers); $s\%$ update sparsity
\Require local minibatch size $B$; momentum $m$; learning rate $\eta$; local dataset $\mathcal{X}_k$
\State $u_0^k \gets 0$; $v_0^k \gets 0$
\For{$t = 0, 1, 2,...$}
\State $b \gets$ (sample $B$ data samples from $\mathcal{X}_k$)
\State $g_{t+1}^k \gets - \eta \cdot \bigtriangledown f(w_t, b)$
\State $g_{t+1}^k \gets \texttt{gradient\_clipping}(g_{t+1}^k)$
\Comment{Clip gradients}
\State $u_{t+1}^k \gets m \cdot u_t^k + g_{t+1}^k$
\State $v_{t+1}^k \gets v_{t}^k + u_{t+1}^k$ 
\Comment{Accumulate weight updates}
\State $T \gets s\% \text{ of } ||v_{t+1}^k||$ 
\Comment{Determine the threshold for sparsified updates}
\For{$j = 0, 1,...M$}
\State $S \gets ||v_{t+1}^k|| > T$
\Comment{Check if accumulated updates are top $s\%$}
\State $\widetilde{v}_{t+1}^k[j] \gets v_{t+1}^k[j] \odot S$
\Comment{Share top updates with other $P_k$}
\State $v_{t+1}^k[j] \gets v_{t+1}^k[j] \odot \neg S$
\Comment{Clear top updates locally}
\State $u_{t+1}^k[j] \gets u_{t+1}^k[j] \odot \neg S$
\Comment{Clear the history of top updates (momentum correction)}
\EndFor
\State $w_{t+1} = w_t + \sum_{k=1}^K \widetilde{v}_{t+1}^k$
\Comment{Apply top updates from all $P_k$}
\EndFor
\end{algorithmic}
\end{algorithm}


\section{Details of Geographical Distribution of Mammal Pictures on Flickr}
\label{appendix:dataset}

\subsection{Dataset Details} 
We query Flickr for the top 40,000 images (4000 images from each of 10 years) for each of the 48 mammal classes in Open Images V4~\cite{DBLP:journals/corr/abs-1811-00982}. We then use PNAS~\cite{DBLP:conf/eccv/LiuZNSHLFYHM18} to clean the search results. 
As PNAS is pre-trained on ImageNet, we can only consider classes that exist both in Open Image and ImageNet. 
As a result, we remove 7 classes from our dataset (Bat, Dog, Raccoon, Giraffe, Rhinoceros, Horse, Mouse). 
Note that while ImageNet has many dogs, they are categorized into hundreds of classes. Hence, we remove dogs in our dataset for simplicity. 
We run all the images through PNAS, and keep all the images with a matching class result in the top-5 predictions.

\begin{figure}[h]
\centering
\includegraphics[width=0.98\textwidth]{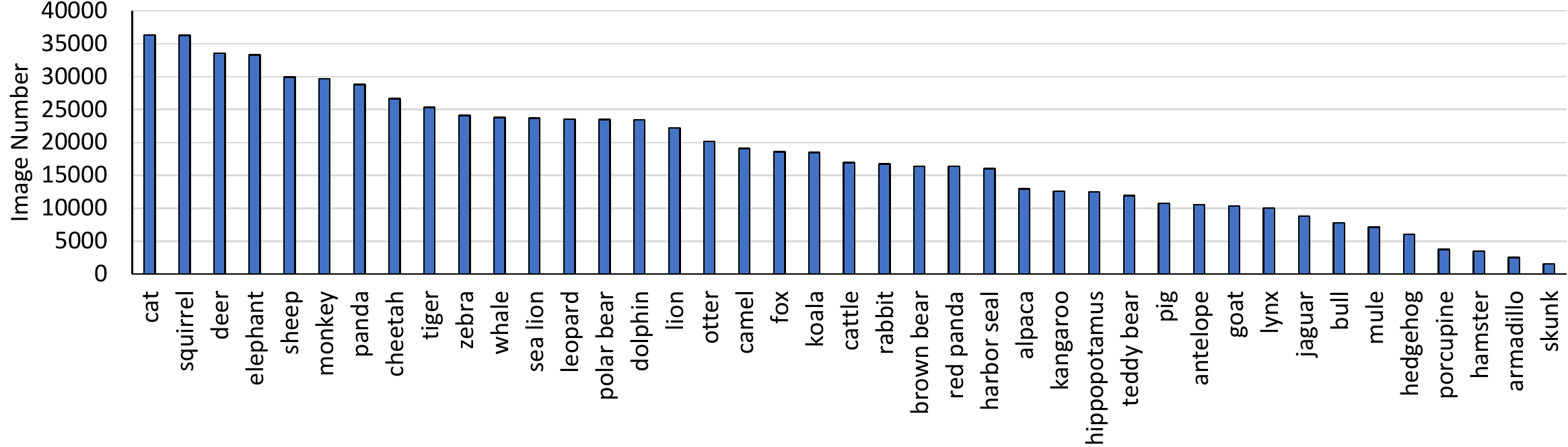}
\caption{\FlickrMammal dataset: The number of images in each mammal class.} 
\label{fig:geo_animal_image_num}
\end{figure}

Figure~\ref{fig:geo_animal_image_num} shows the number of images in each class of our \FlickrMammal dataset. 
As expected, popular mammals (e.g., cat and squirrel) have a lot more images than less popular mammals (e.g., armadillo and skunk). 
The gap between different classes is large: the most popular mammal (cat) has 23$\times$ more images than the least popular mammal (skunk). 
Nonetheless, the vast majority of classes have at least 10,000 images. 
Even the least popular mammal has 1,531 images, which is a reasonable number for DNN training. 
In comparison, ImageNet Large Scale Visual Recognition Challenge (ILSVRC) 2014 has around 1,200 images for each class.

\subsection{First-Level Geographical Region Analysis}

\begin{figure}[h]
\centering
\includegraphics[width=0.98\textwidth]{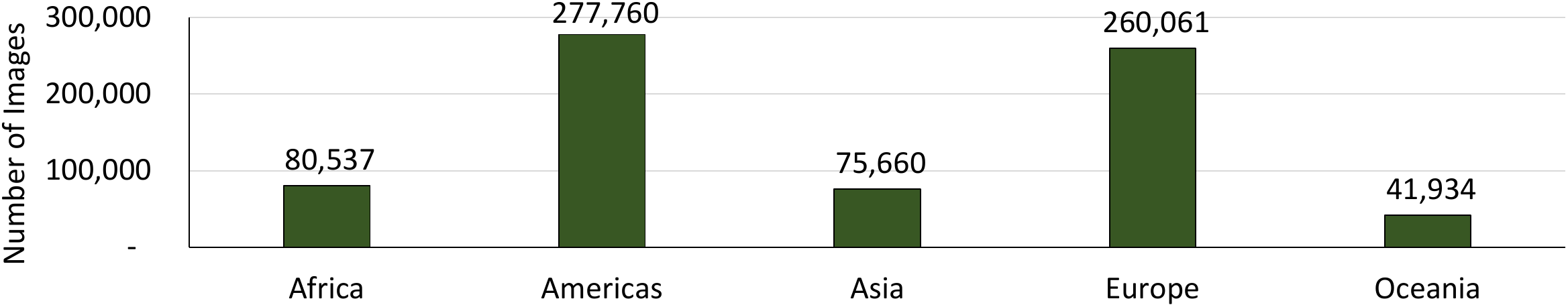}
\caption{\FlickrMammal dataset: The number of images in each continent.} 
\label{fig:geo_animal_first_level_raw}
\end{figure}

As \xref{subsec:dataset} mentions, we use the M49 Standard~\cite{united2019methodology} to map the geotag of each image to different regions. The first-level regions in the M49 Standard are the continents. \pg{Figure~\ref{fig:geo_animal_first_level_raw} shows the number of images in each continent (our analysis omits the 53 images that were not from any one of these five continents).  
There is an inherent skew in the number of images in each continent:}
Americas and Europe have significantly more pictures than the other continents, probably because these two continents have more people who use Flickr to upload pictures.

\begin{figure}[h]
\centering
\includegraphics[width=0.98\textwidth]{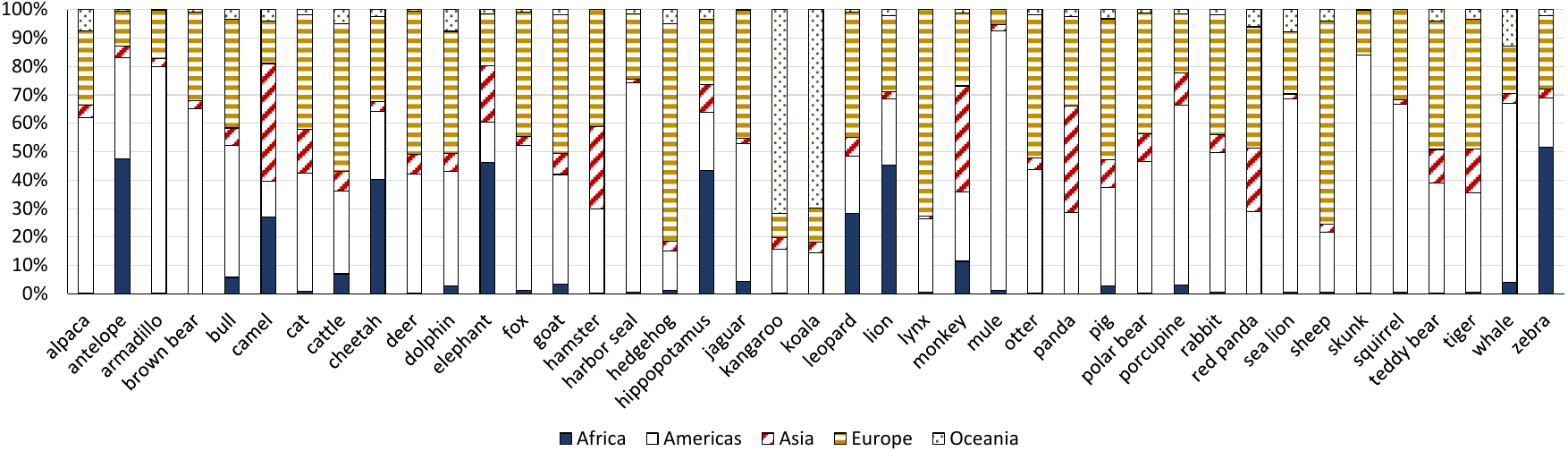}
\caption{\FlickrMammal dataset: The share of images in each continent based on raw samples.} 
\label{fig:geo_animal_first_level_raw_share}
\end{figure}

\textbf{Share of raw samples across continents.}
Figure~\ref{fig:geo_animal_first_level_raw_share} depicts the share of samples across continents for each mammal class. 
As expected, Americas and Europe dominate the share of images for many mammals as they have more images than other continents (Figure~\ref{fig:geo_animal_first_level_raw}). 
However, the geographical distribution of mammals is the main reason for the skew in the share distribution. 
For example, Oceania has more than 70\% of Kangaroo and Koala images even though it only has 6\% of the total images. 
Similarly, Africa has more than 40\% of Antelope, Cheetah, Elephant, Hippopotamus, Lion, and Zebra images while it has only 11\% of the total images.
Overall, we see that the vast majority of mammals are dominated by two or three continents, leaving the other continents with a small number of image samples for these mammal classes. 

\begin{figure}[h]
\centering
\includegraphics[width=0.98\textwidth]{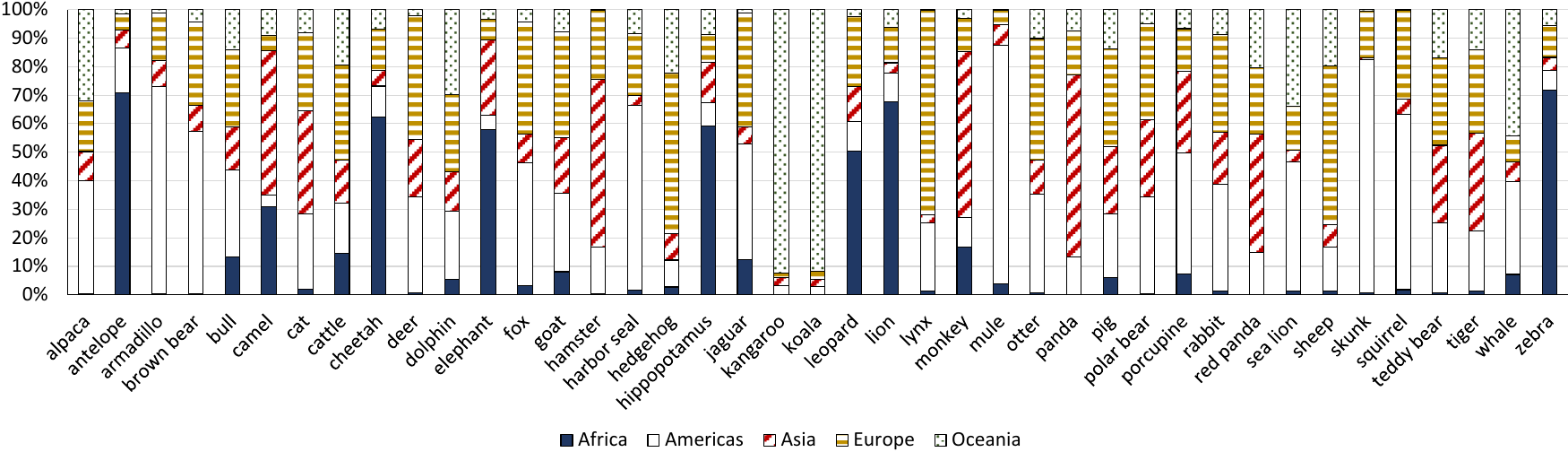}
\caption{\FlickrMammal dataset: The share of images in each continent based on normalized samples.} 
\label{fig:geo_animal_first_level_normalized_share}
\end{figure}

\textbf{Share of normalized samples across continents.}
As we are mostly interested in the distribution of labels ($\mathcal{P}(y)$) among different continents, we normalize the number of images so that each continent has \emph{the same number of total images}. 
Table~\ref{tbl:top_animal_continent} in \xref{subsec:dataset} shows the top-5 mammals in each continent based on \pg{these} normalized samples.
Here, Figure~\ref{fig:geo_animal_first_level_normalized_share} illustrates the normalized sample share for all mammals across \pg{continents}.
As we see, the overall label distribution is similar between normalized \pg{samples} (Figure~\ref{fig:geo_animal_first_level_normalized_share}) and raw samples (Figure~\ref{fig:geo_animal_first_level_raw_share}).  
The continent that dominates a mammal class in the raw sample distribution tends to be even more dominant in the normalized sample distribution. 
For example, Africa consists of 50\% to 70\% of the African mammals (e.g., Antelope, Cheetah, Elephant, etc.) in the normalized sample distribution, compared to 40\% in the raw sample distribution. 
We conclude that skewed distribution of labels is a natural phenomenon, and both raw samples and normalized samples exhibit very significant skew across common mammals.

\subsection{Second-Level Geographical Region Analysis}
\label{subsec:second_level_mammal}

\begin{figure}[h]
\centering
\includegraphics[width=0.98\textwidth]{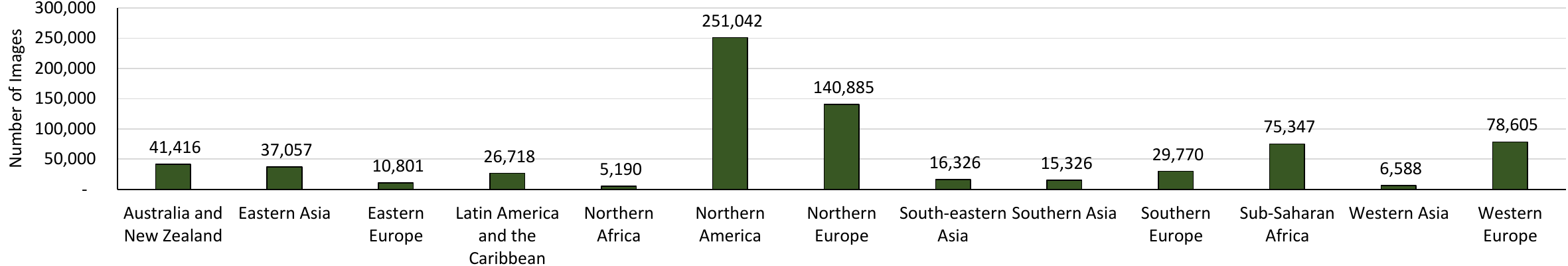}
\vspace{-0.1in}
\caption{\FlickrMammal dataset: The number of images in each subcontinent.} 
\label{fig:geo_animal_second_level_raw}
\vspace{-0.05in}
\end{figure}

We also analyze our dataset using the second-level regions (subcontinents) in the M49 Standard. 
We remove the second-level regions that have fewer than 1,000 images in our analysis (Central Asia, Melanesia, Micronesia, and Polynesia), \pg{resulting in 13 subcontinents and 735,071 images}.
Figure~\ref{fig:geo_animal_second_level_raw} shows the number of images in each subcontinent. 
Similar to Figure~\ref{fig:geo_animal_first_level_raw}, we see that Northern America and Northern Europe have significantly more images than other subcontinents.

\begin{figure}[h]
\centering
\includegraphics[width=0.98\textwidth]{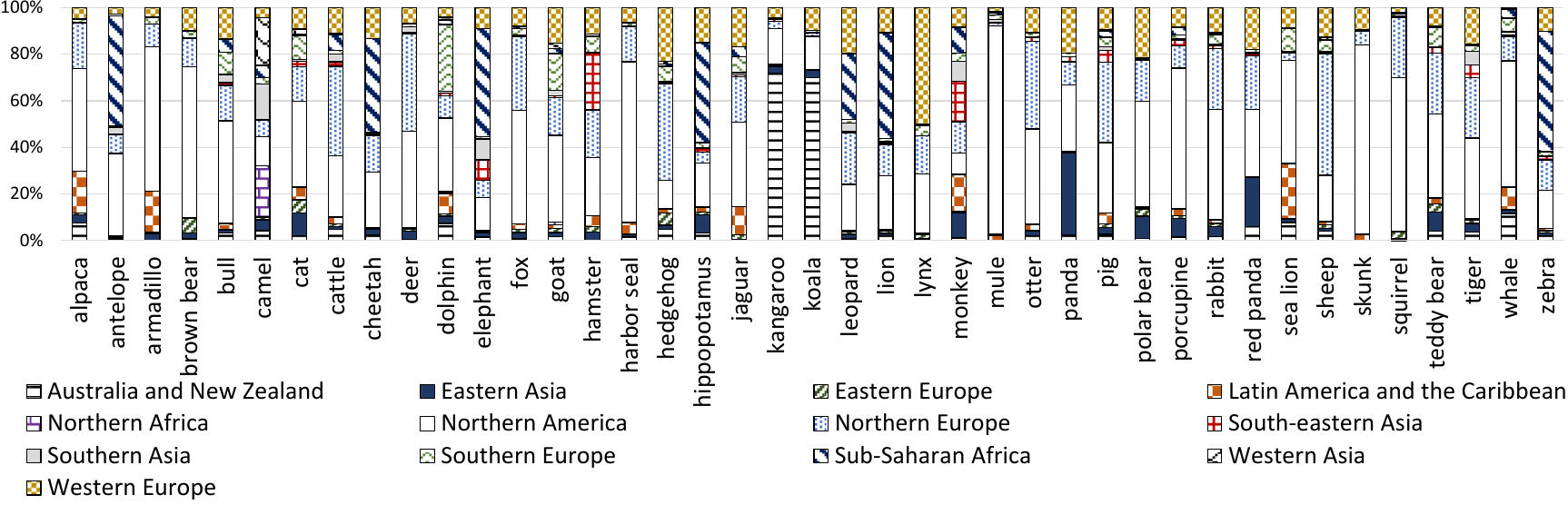}
\vspace{-0.1in}
\caption{\FlickrMammal dataset: The share of images in each subcontinent based on raw samples.} 
\label{fig:geo_animal_first_second_raw_share}
\vspace{-0.12in}
\end{figure}

\begin{figure}[h]
\centering
\includegraphics[width=0.98\textwidth]{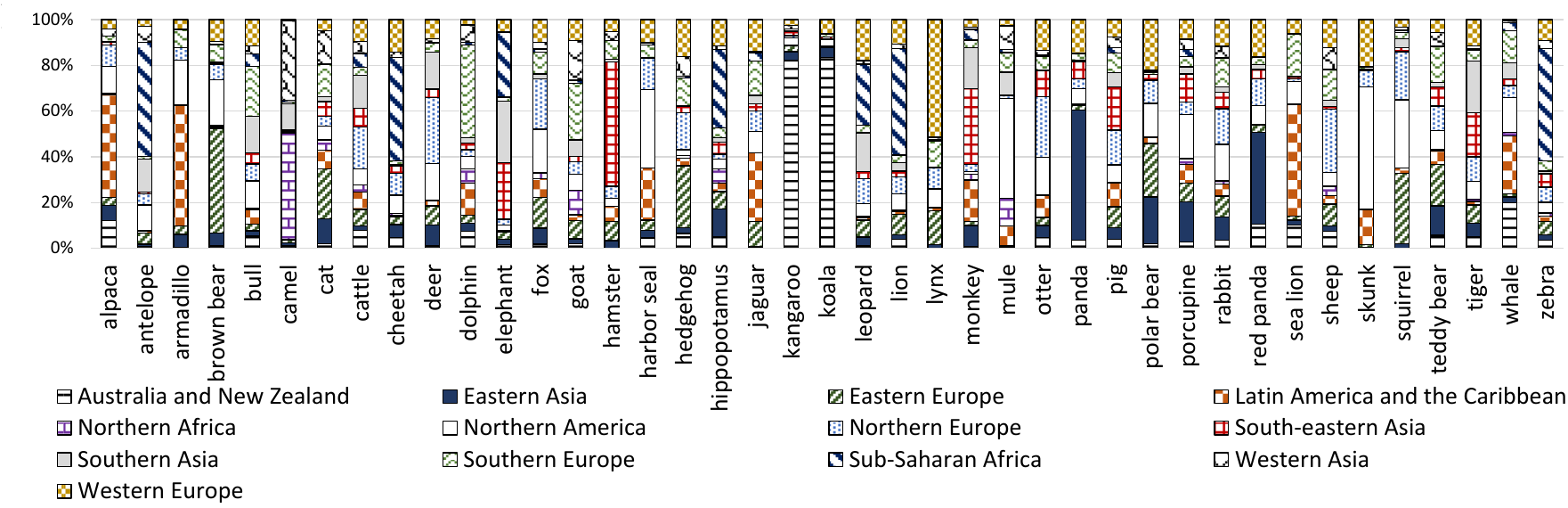}
\vspace{-0.1in}
\caption{\FlickrMammal dataset: The share of images in each subcontinent based on normalized samples.} 
\label{fig:geo_animal_second_level_normalized_share}
\end{figure}
\clearpage

\textbf{Share of samples across subcontinents.}
Figure~\ref{fig:geo_animal_first_second_raw_share} illustrates the share of samples across subcontinents for each mammal class. Again, we observe that the label distribution is highly skewed. Among the 13 subcontinents, the vast majority of mammal classes mostly exist in 3-5 subcontinents. Furthermore, the sample concentration pattern varies greatly among mammal classes. For example, Kangaroo and Koala are mostly in Australia and New Zealand, \pg{Antelope and Zebra} 
are mostly in Sub-Saharan Africa, and Mule and Skunk are mostly in Northern America. On average, 5 of the 13 subcontinents contain less than 1\% of \pg{the} images for each mammal class. We also show the normalized sample share across subcontinents (Figure~\ref{fig:geo_animal_second_level_normalized_share}), and we can see the difference of $\mathcal{P}(y)$ among subcontinents. Overall, our analysis shows that skewed label distribution is also very common at the subcontinent-level.

\section{Training Parameters}
\label{appendix:training_parameter}

Tables~\ref{tbl:training_parameter_cifar10}--\ref{tbl:training_parameter_flickr} list the major training parameters for all the applications, models, and datasets in our study.

\begin{table}[h!]
  \centering
  \small
  \begin{tabular}{c c c c c c}
    \toprule
    \textbf{Model} & \scell{\textbf{Minibatch size} \\\textbf{\ \ \ \ \ per node} \\ \textbf{\ \ \ \ \ (5 nodes)}} & \textbf{Momentum} & \scell{\textbf{Weight}\\ \textbf{\, decay}} & \textbf{Learning rate} & \textbf{Total epochs} \\ \midrule
    AlexNet & 20 & 0.9 & 0.0005 & \scell{$\eta_0 = 0.0002$, divides by \\10 at epoch 64 and 96} & 128 \\ \midrule
    GoogLeNet & 20 & 0.9 & 0.0005 & \scell{$\eta_0 = 0.002$, divides by \\10 at epoch 64 and 96} & 128 \\ \midrule
    \scell{LeNet, BN-LeNet,\\GN-LeNet} & 20 & 0.9 & 0.0005 & \scell{$\eta_0 = 0.002$, divides by \\10 at epoch 64 and 96} & 128 \\ \midrule
    ResNet-20 & 20 & 0.9 & 0.0005 & \scell{$\eta_0 = 0.002$, divides by \\10 at epoch 64 and 96} & 128 \\ 
    \bottomrule \\
  \end{tabular}
  \vspace{-10pt}
  \caption{Major training parameters for \appimage over CIFAR-10}
  \label{tbl:training_parameter_cifar10}
\end{table}

\begin{table}[h!]
  \centering
  \small
  \begin{tabular}{c c c c c c}
    \toprule
    \textbf{Model} & \scell{\textbf{Minibatch size} \\\textbf{\ \ \ \ \ per node} \\ \textbf{\ \ \ \ \ (8 nodes)}} & \textbf{Momentum} & \scell{\textbf{Weight}\\ \textbf{\, decay}} & \textbf{Learning rate} & \textbf{Total epochs} \\ \midrule
    GoogLeNet & 32 & 0.9 & 0.0002 & \scell{$\eta_0 = 0.0025$, polynomial \\decay, power = 0.5} & 60 \\ \midrule
    ResNet-10 & 32 & 0.9 & 0.0001 & \scell{$\eta_0 = 0.00125$, polynomial \\decay, power = 1} & 64 \\ 
    \bottomrule \\
  \end{tabular}
  \vspace{-10pt}
  \caption{Major training parameters for \appimage over ImageNet. Polynomial decay means $\eta = \eta_0 \cdot (1 - \frac{\text{iter}}{\text{max\_iter}})^{\text{power}}$.}
  \label{tbl:training_parameter_imagenet}
\end{table}

\begin{table}[h!]
  \centering
  \small
  \begin{tabular}{c c c c c c}
    \toprule
    \textbf{Model} & \scell{\textbf{Minibatch size} \\\textbf{\ \ \ \ \ per node} \\ \textbf{\ \ \ \ \ (4 nodes)}} & \textbf{Momentum} & \scell{\textbf{Weight}\\ \textbf{\, decay}} & \textbf{Learning rate} & \textbf{Total epochs} \\ \midrule
    center-loss & 64 & 0.9 & 0.0005 & \scell{$\eta_0 = 0.025$, divides by \\10 at epoch 4 and 6} & 7 \\ 
    \bottomrule \\
  \end{tabular}
  \vspace{-5pt}
  \caption{Major training parameters for \appface over CASIA-WebFace.}
  \label{tbl:training_parameter_face}
\end{table}

\begin{table}[h!]
  \centering
  \small
  \begin{tabular}{c c c c c c}
    \toprule
    \textbf{Model} & \scell{\textbf{Minibatch size} \\\textbf{\ \ \ \ \ per node} \\ \textbf{\ \ \ \ \ (5 nodes)}} & \textbf{Momentum} & \scell{\textbf{Weight}\\ \textbf{\, decay}} & \textbf{Learning rate} & \textbf{Total epochs} \\ \midrule
    GoogLeNet & 32 & 0.9 & 0.0002 & \scell{$\eta_0 = 0.004$, polynomial \\decay, power = 0.5} & 55 \\ 
    \bottomrule \\
  \end{tabular}
  \vspace{-5pt}
  \caption{Major training parameters for \appimage over \FlickrMammal.}
  \label{tbl:training_parameter_flickr}
\end{table}
\clearpage

{\kvb

\section{Training Convergence Curves}
\label{appendix:convergence_curves}

Figures \ref{fig:cifar10_alexnet_curve} and \ref{fig:cifar10_resnet_curve} show the training convergence curves for AlexNet and ResNet20 over the CIFAR-10 dataset.
We make two major observations. 
First, all training processes stop improving long before the end of experiments, which suggest longer training cannot solve the problem of non-IID data. 
Second, the convergence curves in the Non-IID settings generally follow similar trends to the curves in the IID settings, but the model accuracy is significantly lower. Appendix \ref{sec:decentral_cause} discusses the potential reasons behind this phenomenon. 
As we discuss in \xref{sec:batch_norm}, even BSP loses significant accuracy for DNN models with BatchNorm, which explains the curves in Figure \ref{fig:cifar10_resnet_curve}.  

\begin{figure}[h!]
  \centering
  \begin{subfigure}[t]{0.27\linewidth}
    \centering
    \includegraphics[width=1.0\textwidth]{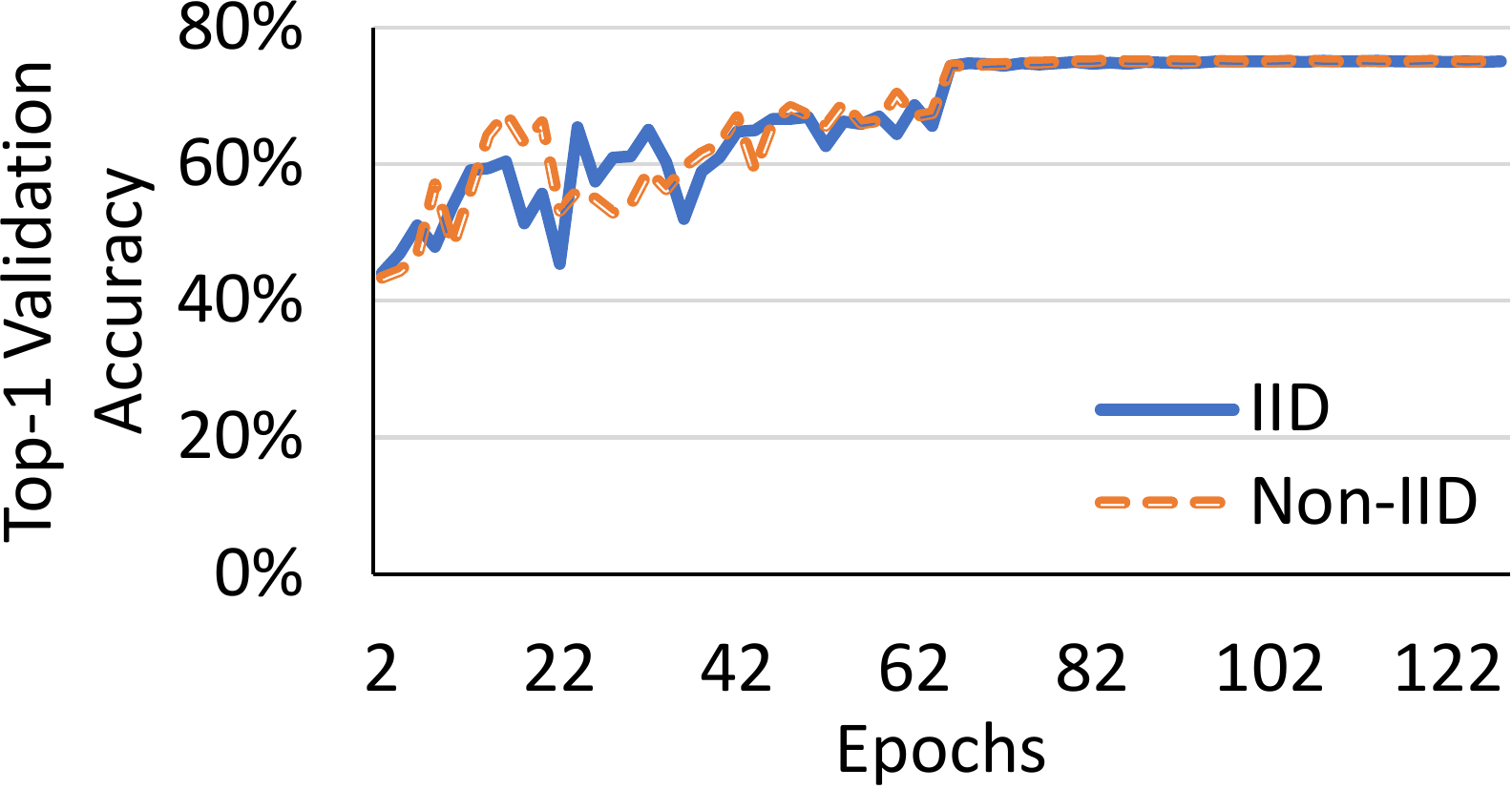}
    \caption{BSP}
  \end{subfigure}
   \centering 
  \begin{subfigure}[t]{0.23\linewidth}
    \centering
    \includegraphics[width=1.0\textwidth]{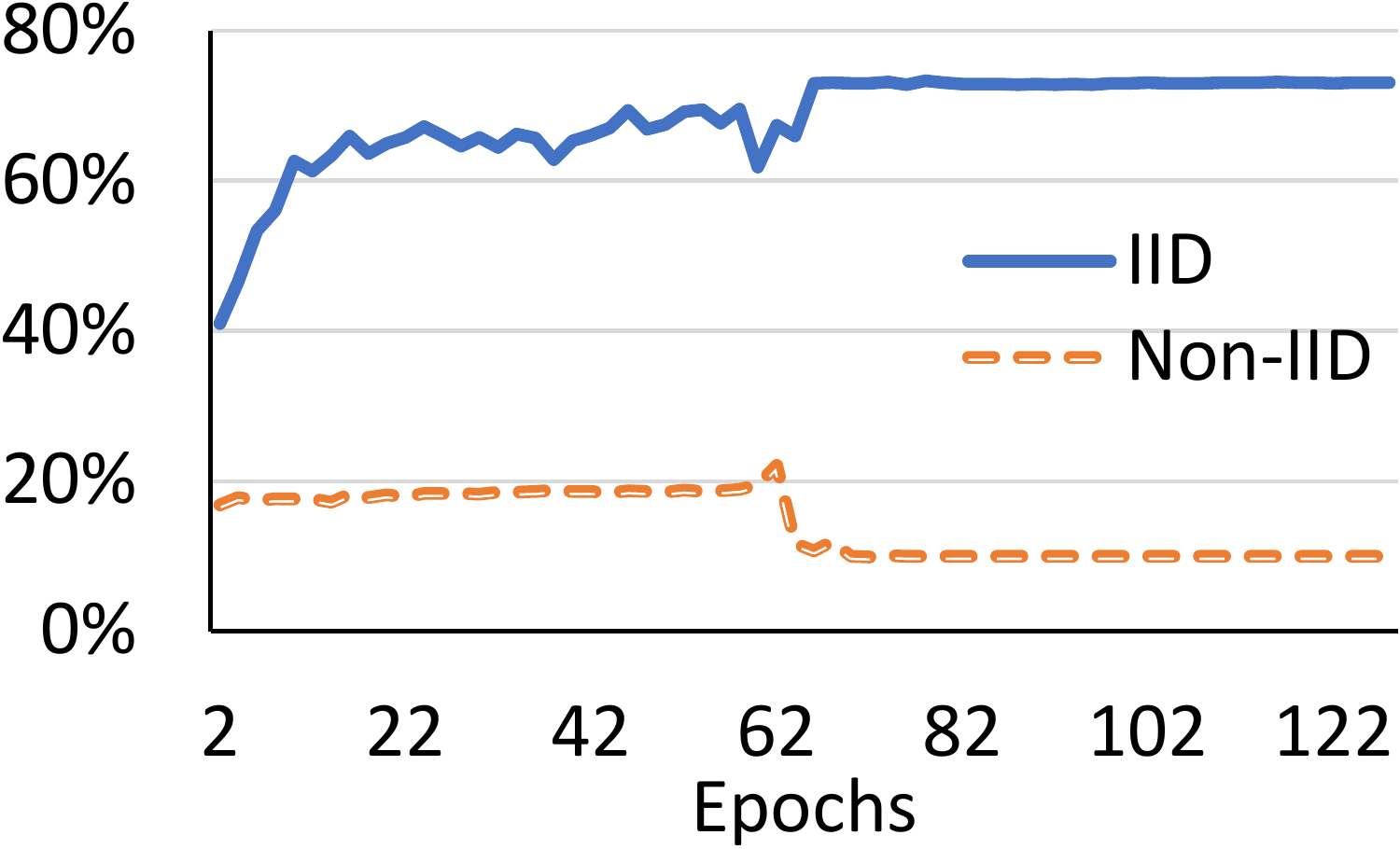}
    \caption{Gaia}
  \end{subfigure}
  \begin{subfigure}[t]{0.23\linewidth}
    \centering
    \includegraphics[width=1.0\textwidth]{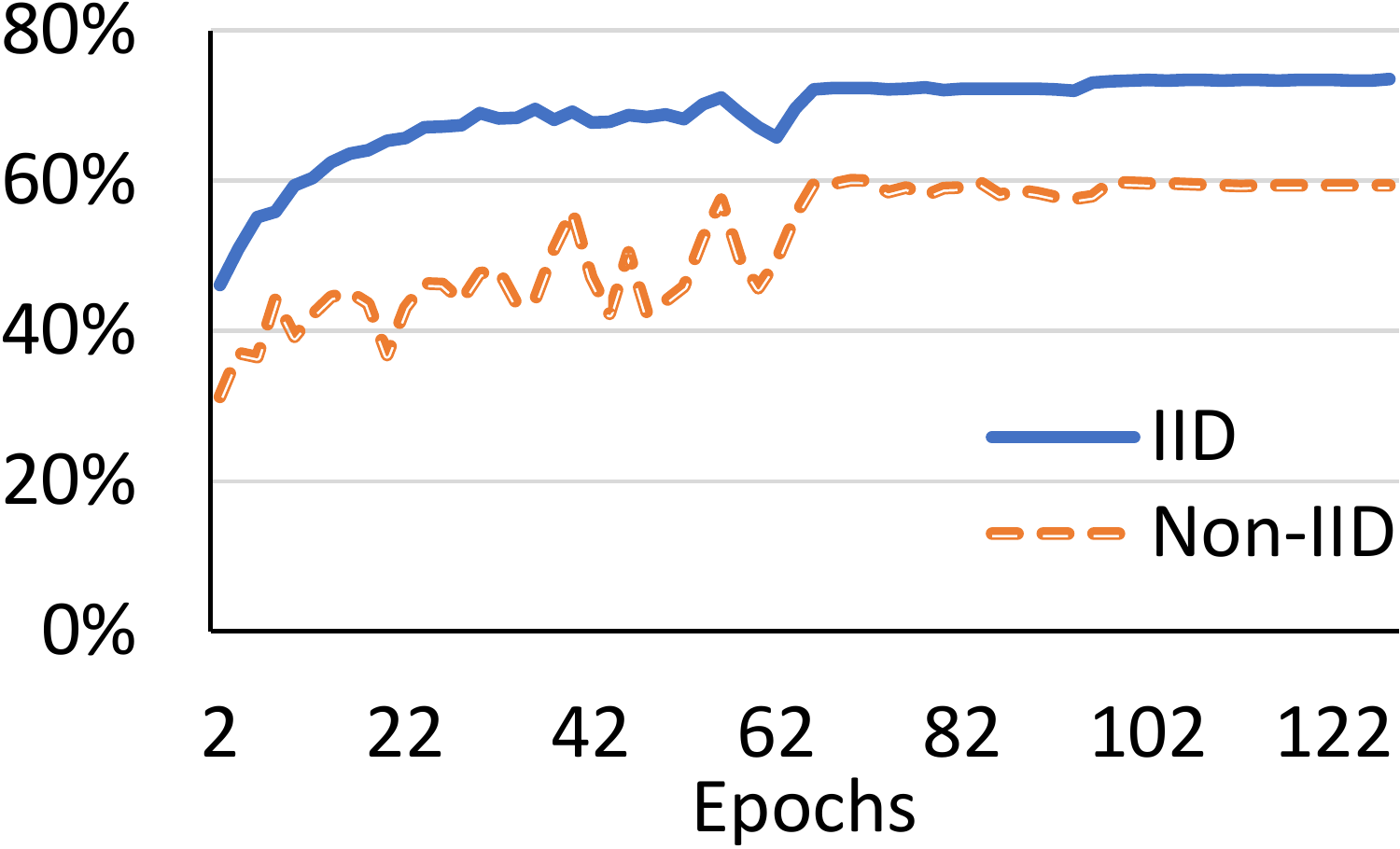}
    \caption{Federated Averaging}
  \end{subfigure}
  \begin{subfigure}[t]{0.23\linewidth}
    \centering
    \includegraphics[width=1.0\textwidth]{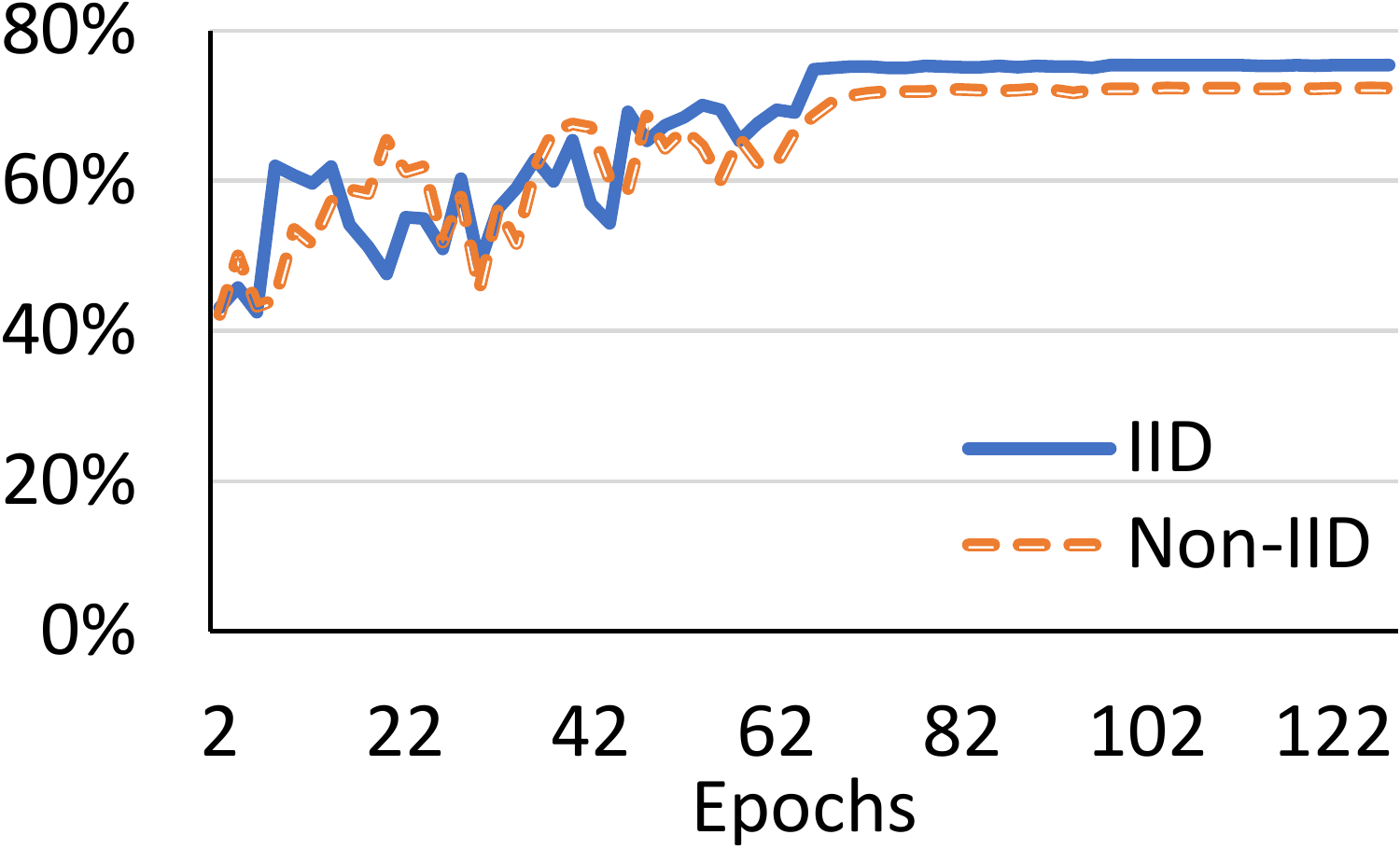}
    \caption{Deep Gradient Compression}
  \end{subfigure}
    \vspace{-0.05in}
  \caption{The training convergence curves for AlexNet over the CIFAR-10 dataset.}
  \label{fig:cifar10_alexnet_curve}
\end{figure}

\begin{figure}[h!]
  \centering
  \begin{subfigure}[h]{0.27\linewidth}
    \centering
    \includegraphics[width=1.0\textwidth]{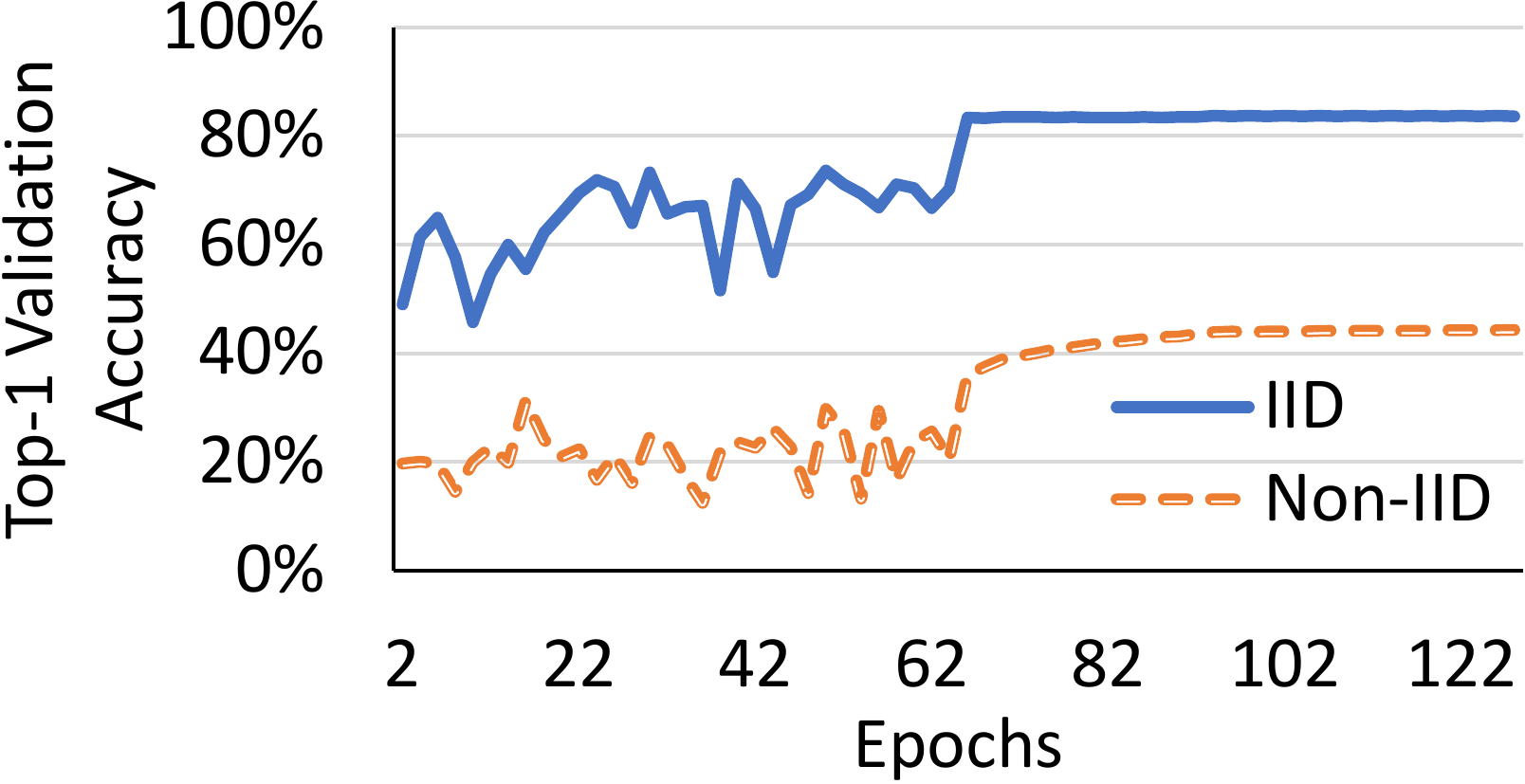}
    \caption{BSP}
  \end{subfigure}
   \centering 
  \begin{subfigure}[h]{0.23\linewidth}
    \centering
    \includegraphics[width=1.0\textwidth]{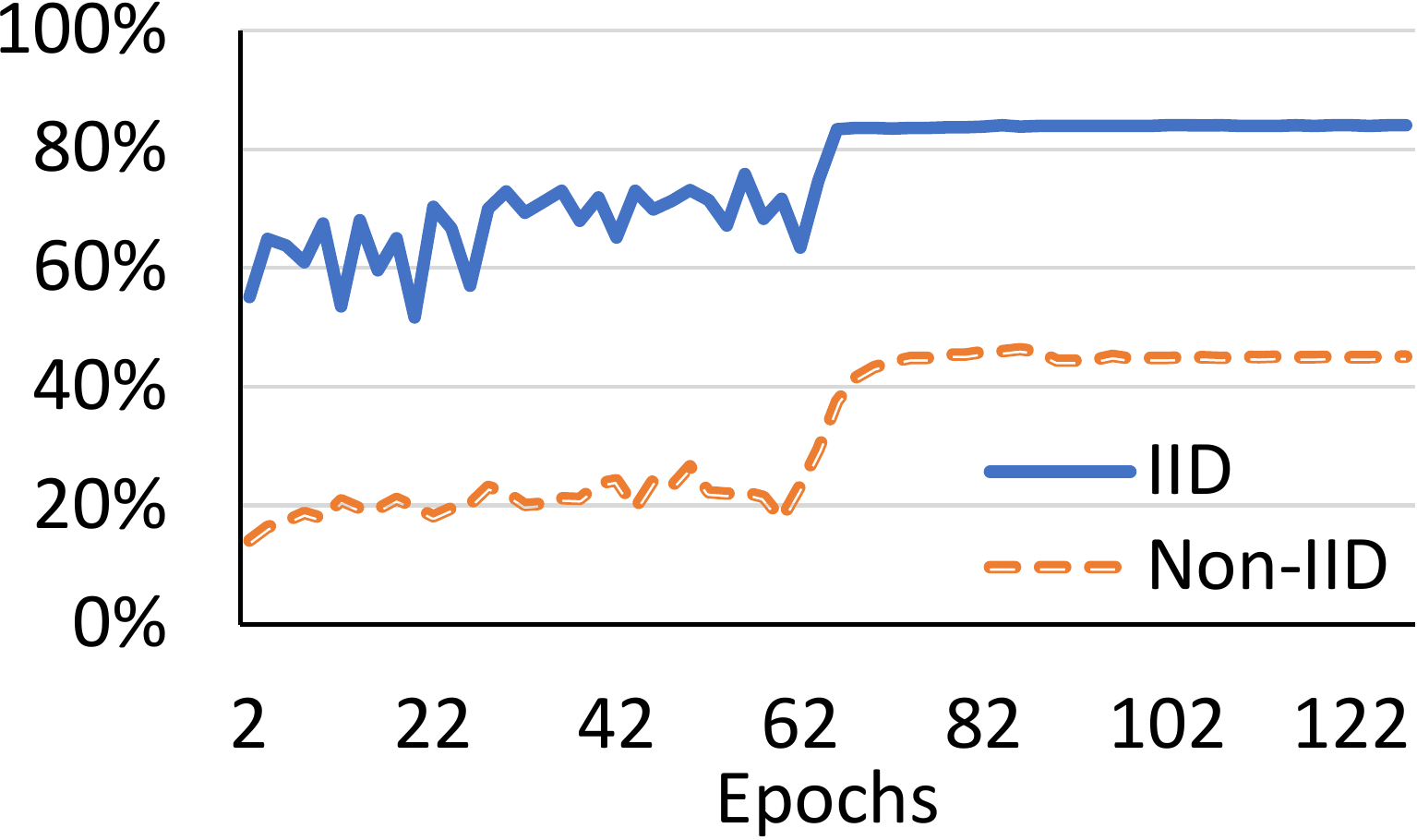}
    \caption{Gaia}
  \end{subfigure}
  \begin{subfigure}[h]{0.23\linewidth}
    \centering
    \includegraphics[width=1.0\textwidth]{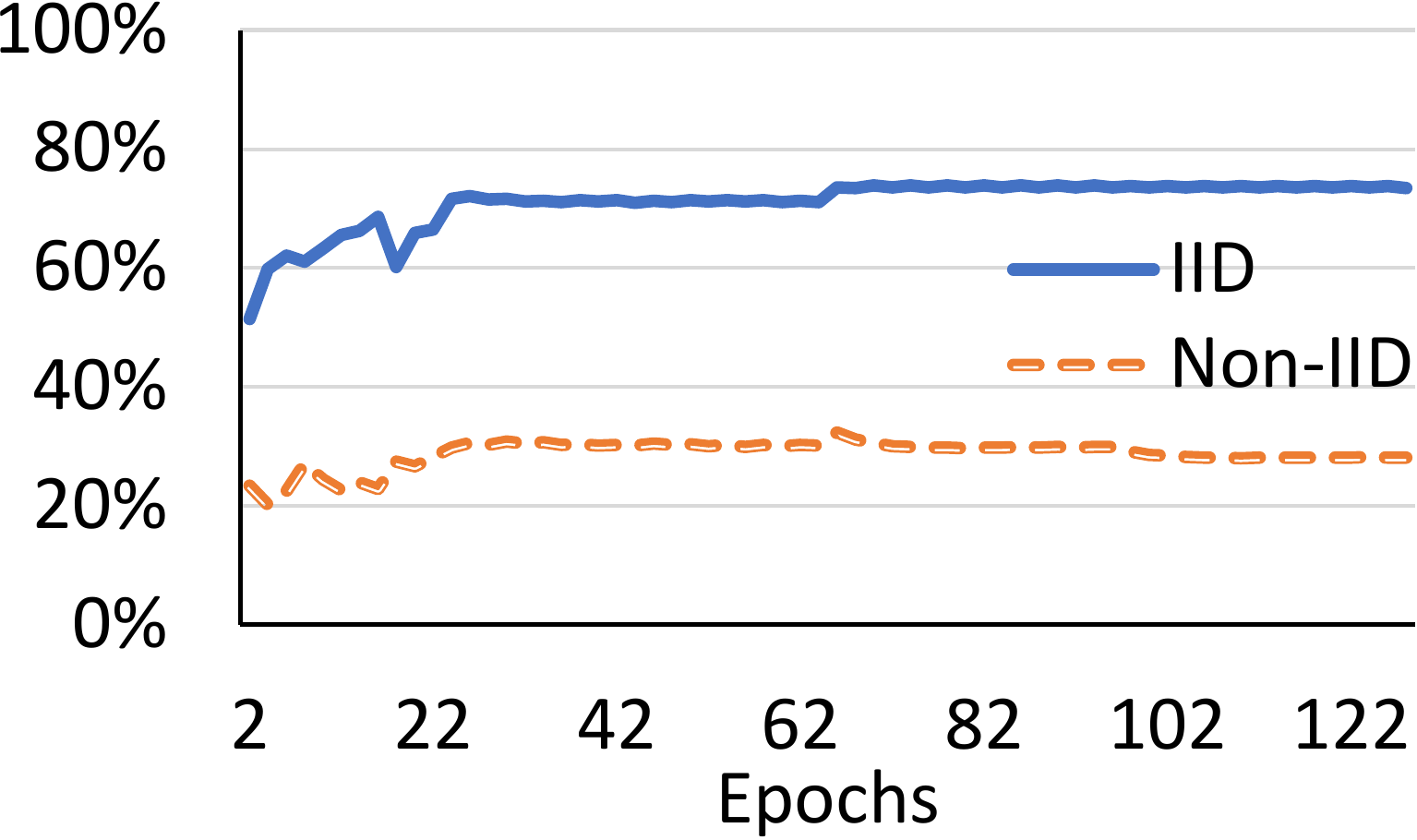}
    \caption{Federated Averaging}
  \end{subfigure}
  \begin{subfigure}[h]{0.23\linewidth}
    \centering
    \includegraphics[width=1.0\textwidth]{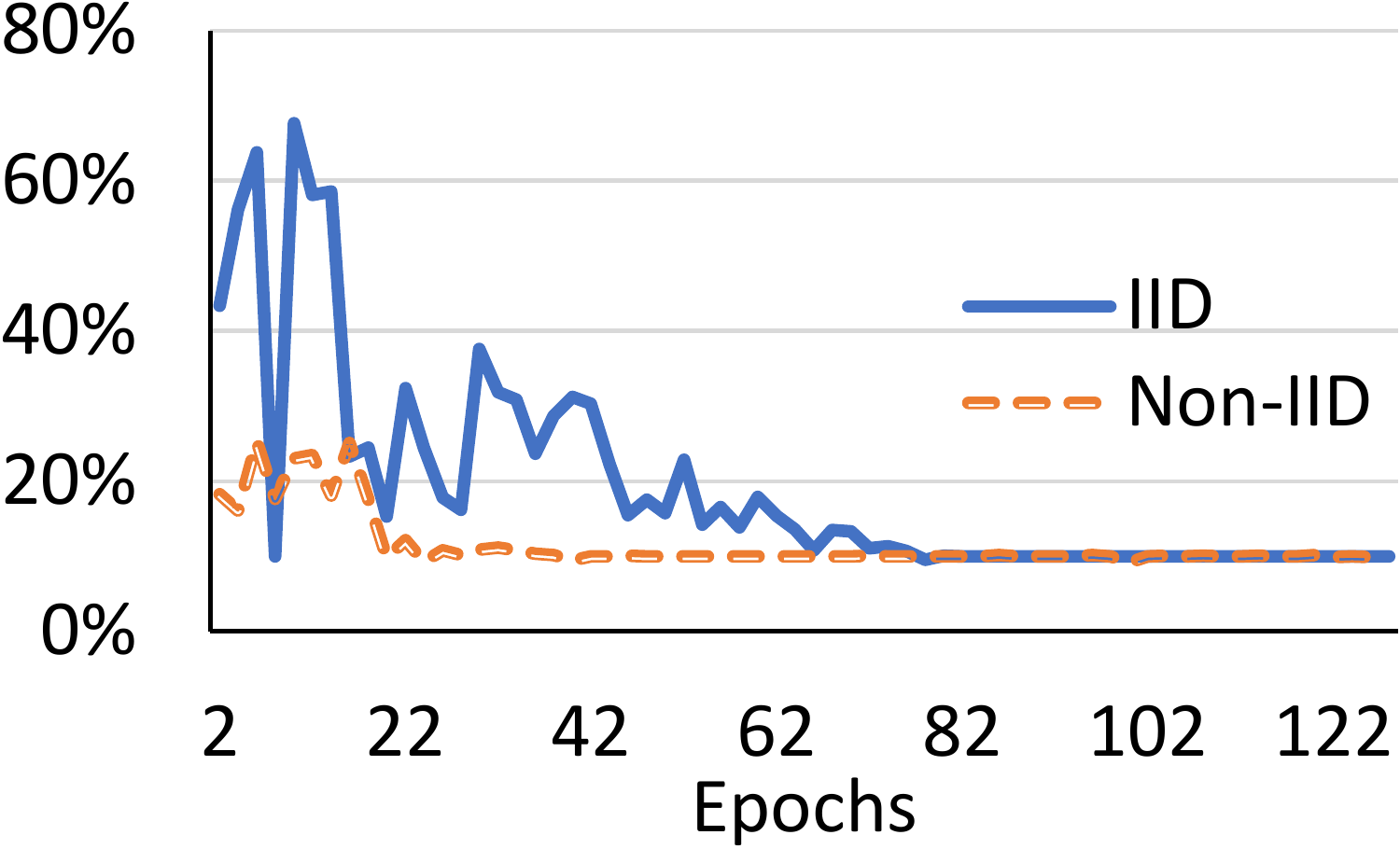}
    \caption{Deep Gradient Compression}
  \end{subfigure}
  \vspace{-0.05in}
  \caption{The training convergence curves for ResNet20 over the CIFAR-10 dataset.}
  \label{fig:cifar10_resnet_curve}
\end{figure}
}

\section{Image Classification with ImageNet}
\label{sec:appendix_imagenet}

\xref{subsec:overview_cifar10} summarized our results for \appimage over the ImageNet dataset~\cite{ILSVRC15} (1,000 image classes).
In this section, we provide the details.

\begin{figure}[h!]
\centering
\includegraphics[width=0.55\textwidth]{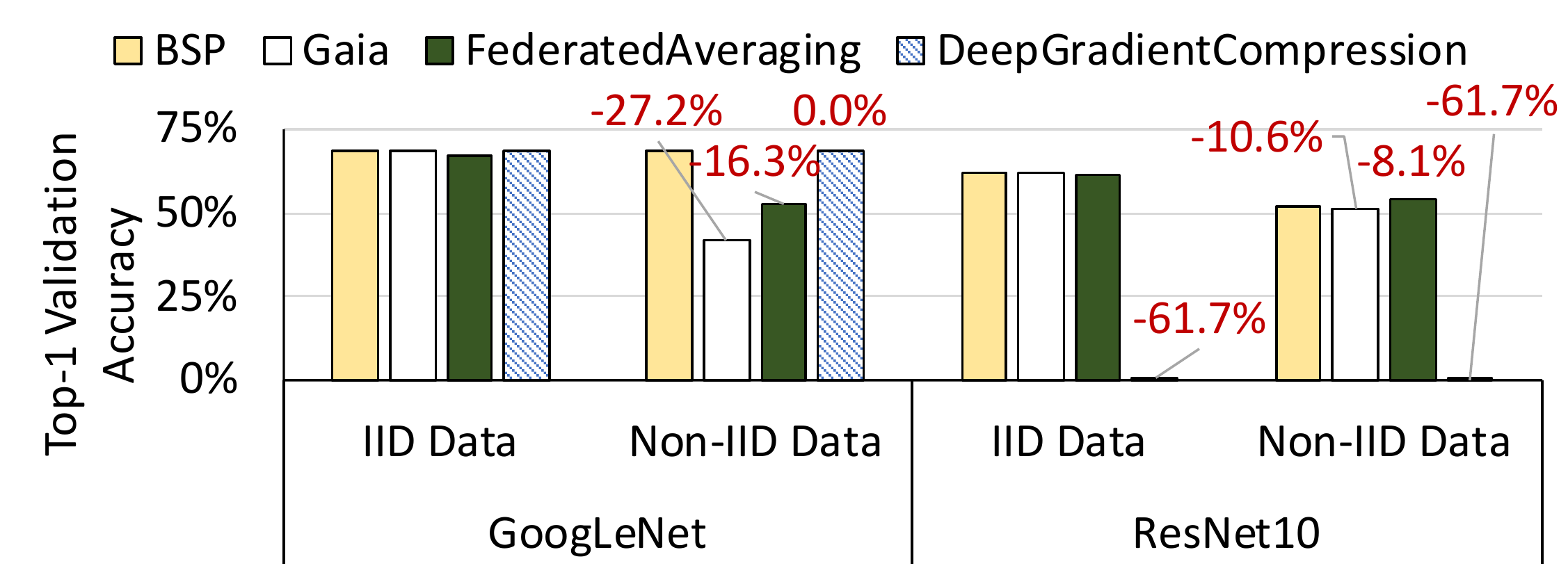}\vspace{-0.15in}
\caption{Top-1 validation accuracy for \appimage over the ImageNet dataset. Each ``-x\%'' label indicates the accuracy loss \pg{relative to} BSP in the IID setting.} 
\label{fig:overview_imagenet}
\end{figure}

We use two partitions ($K = 2$) in this experiment so each partition contains 500 image classes. 
According to the hyper-parameter criteria in \xref{sec:setup}, we select $T_0 = 40\%$ for \gaia, $Iter_{Local} = 200$ for \fedavg, and $E_{warm} = 4$ for \dgc. \pg{Figure~\ref{fig:overview_imagenet} shows} the validation accuracy in the IID and Non-IID settings.

Interestingly, we observe the same problems in the ImageNet dataset as in the CIFAR-10 dataset (\xref{subsec:overview_cifar10}), even though the number of \pg{classes in ImageNet is two orders of magnitude larger than in CIFAR-10.} 
First, we see that \gaia and \fedavg lose significant validation accuracy (8.1\% to 27.2\%) for both DNNs in the Non-IID setting. 
On the other hand, while \dgc is able to retain the validation accuracy for GoogLeNet in the Non-IID setting, it cannot converge to a useful model for ResNet10. 
Second, BSP also cannot retain the validation accuracy for ResNet10 in the Non-IID setting, which concurs with our observation in the CIFAR-10 study. 
Together with the results in \xref{subsec:overview_cifar10}, these results show that the Non-IID data problem exists not only in various decentralized learning algorithms and DNNs, but also in different \pg{image} datasets.



{\kvb
\section{Effect of Larger Numbers of Data Partitions}
\label{sec:more_partition}

So far, we have used \pg{a relatively} modest number of data partitions \pg{($K=2$ or $K=5$)} to demonstrate the Non-IID data problem in our study. Here, we study the effect of having a larger number of data partitions.

\textbf{CIFAR-10.} We compare the model accuracy of ResNet20 using the CIFAR-10 dataset with ten data partitions ($K=10$). 
We quickly discover that even training with BSP \emph{does not} converge in the 100\% Non-IID setting. 
This is because each partition has only one object class (CIFAR-10 consists of ten object classes), so the gradients from different data partitions diverge too much.
Instead, we create a Non-IID setting such that each partition has 80\% of one object class and 20\% of another object class.
Figure~\ref{fig:cifar10_resnet_more_partition_results} shows the results. The hyper-parameters are the same as in \xref{subsec:overview_cifar10}. 
We observe that decentralized learning algorithms experience similar model accuracy loss with $K=10$ compared to $K=5$.
This is interesting as we have a relatively easier Non-IID setting for $K=10$.
We also observe that with $K=10$, decentralized learning algorithms lose more accuracy relative to BSP in the Non-IID setting compared to $K=5$. When $K=10$, \gaia and \fedavg lose 3\% and 36\% compared to BSP in the Non-IID setting, which is larger than \pg{their} 0\% and 17\% losses when $K=5$.
\pg{These results suggest that} a larger number of data partitions negatively \pg{impacts} model accuracy \pg{in the Non-IID setting}.

\begin{figure}[h!]
\centering
\includegraphics[width=0.70\textwidth]{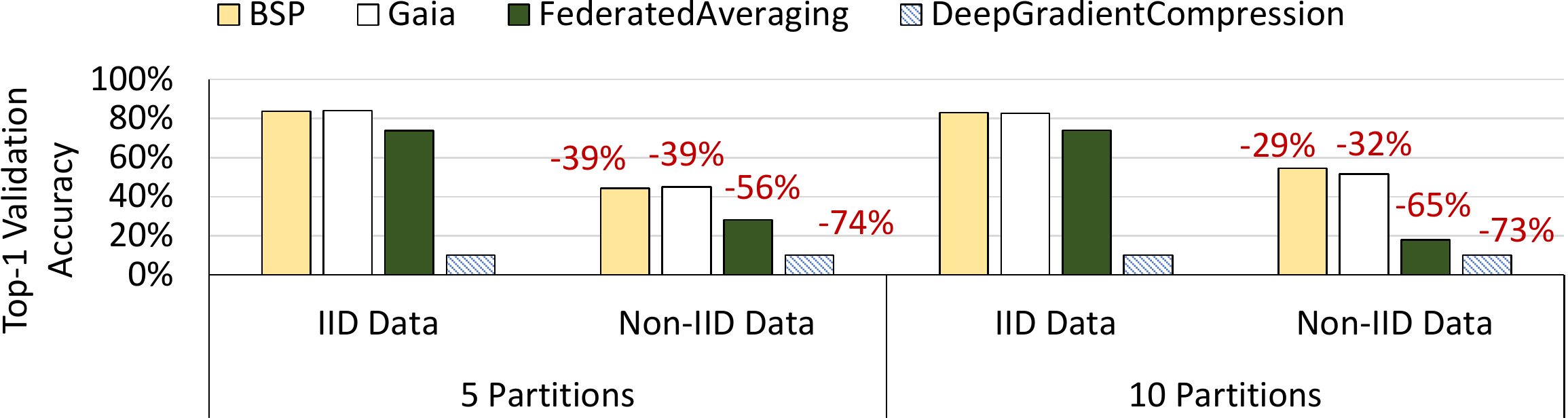}
\vspace{-0.05in}
\caption{\kv{Top-1 validation accuracy for ResNet20 over the CIFAR-10 dataset, with 5 and 10 data partitions. \pg{The five partition results are repeated from Figure~\ref{fig:overview_cifar10}.}
Each ``-x\%'' label indicates the accuracy loss \pg{relative} to BSP in the IID setting.}} 
\label{fig:cifar10_resnet_more_partition_results}
\end{figure}

\textbf{\FlickrMammal.} We create a real-world non-IID setting \pg{using} the locations of mammal images in the second-level (subcontinent) regions. 
As \xref{subsec:second_level_mammal} shows, there are thirteen subcontinents in our dataset so we have $K=13$. 
For comparison, we create an artificial IID setting, in which all images are randomly distributed among the 13 partitions. 
Figure~\ref{fig:second_level_result_geo_animal} shows the results of running BSP, \gaia, and \fedavg in these settings. 
\pg{We use GoogLeNet in this experiment. We select $T_0 = 10\%$ for \gaia and $Iter_{Local} = 20$ for \fedavg based on the criteria in \xref{sec:setup}.}
We see that both \gaia and \fedavg lose more accuracy when data \pg{are} partitioned at the subcontinent level ($K=13$) than at the continent level ($K=5$). 
This is expected because the vast majority of mammals mostly exist in 3-5 subcontinents (Figure~\ref{fig:geo_animal_second_level_normalized_share}), so many subcontinents do not have all the mammal labels. 
In contrast, most continents have all the mammal labels, which reduces the difficulty level of the problem.
This result suggests \pg{that} the non-IID data problem can have a more severe impact with a larger number of data partitions in the real world.

\begin{figure}[h!]
\centering
\includegraphics[width=0.70\textwidth]{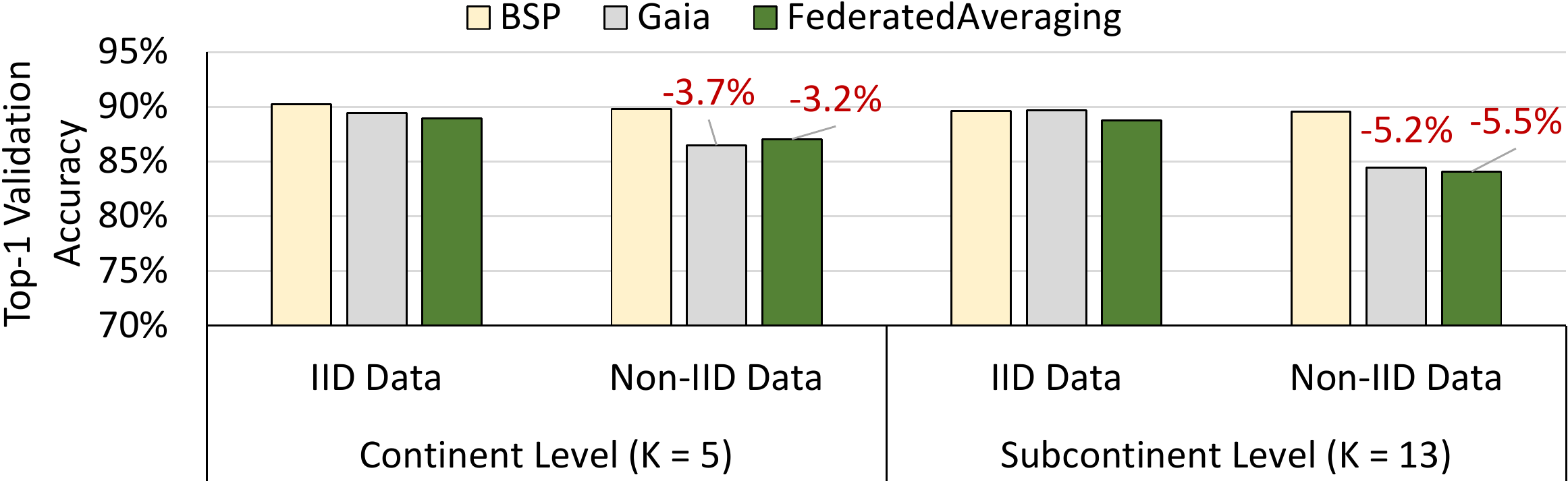}
\vspace{-0.05in}
\caption{\kv{Top-1 validation accuracy for \pg{GoogLeNet} over the \FlickrMammal dataset, which is partitioned at the continent level and the subcontinent level. 5\% of data are randomly selected as the validation set. 
\pg{The five partition results are repeated from Figure~\ref{fig:overview_result_geo_animal}.}
Non-IID Data is based on real-world data distribution among \pg{continents or subcontinents}, and IID Data is the artificial setting in which training images are randomly assigned to partitions.
Each ``-x\%'' label indicates the accuracy loss \pg{relative} to BSP in the IID setting. 
Note: \pg{The} y-axis starts at 70\% accuracy.}} 
\label{fig:second_level_result_geo_animal}
\end{figure}
}
\clearpage

\section{Reasons for Model Quality Loss}
\label{sec:decentral_cause}

\begin{figure}[h]
\centering
\includegraphics[width=0.55\textwidth]{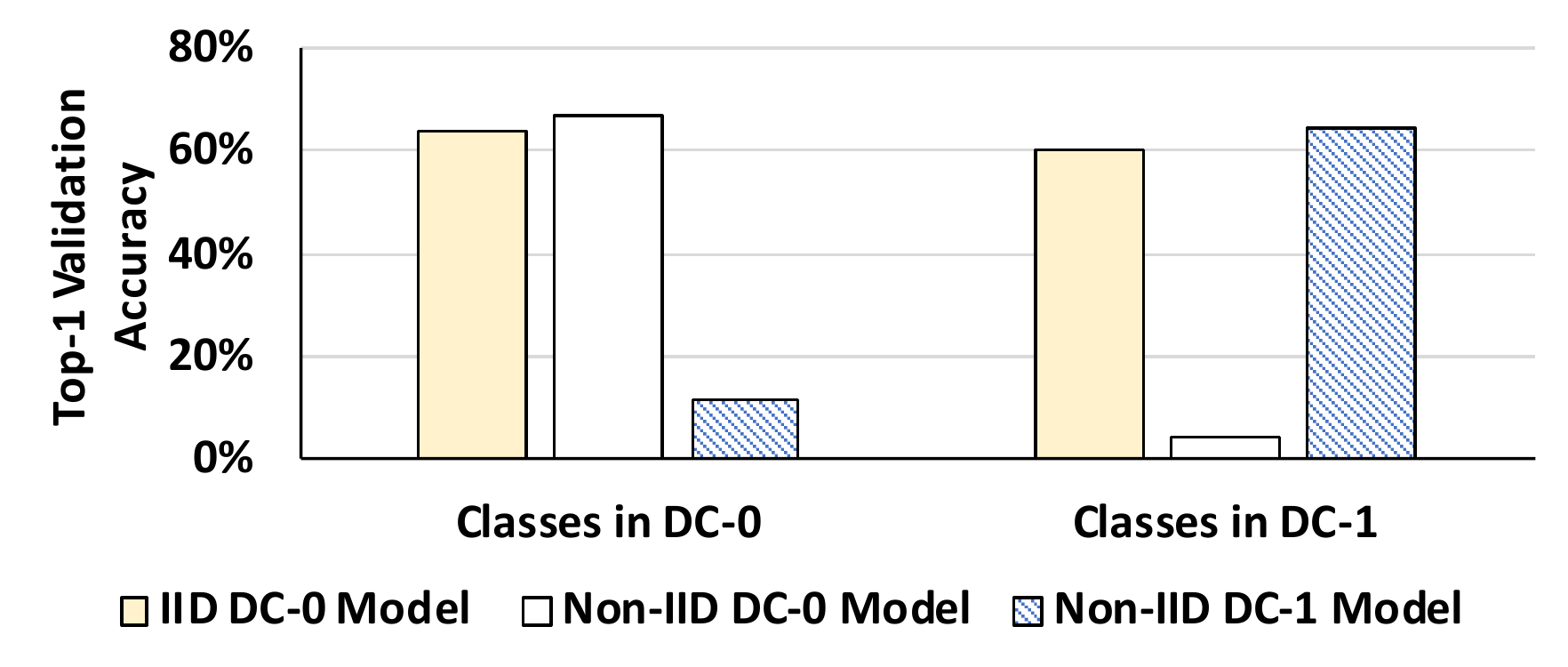}
\vspace{-0.1in}
\caption{Top-1 validation accuracy (ImageNet) for models in different partitions.}
\label{fig:cause_gaia}
\end{figure}

\textbf{Gaia.}
\pg{As discussed in \xref{subsec:cause}, \gaia saves communication by allowing small model differences in each partition $P_k$, and this gives each $P_k$ room for specializing to its local data.
To demonstrate this,} we extract the \gaia-trained models from both partitions (denoted DC-0 and DC-1) \pg{for the GoogLeNet experiment in Figure~\ref{fig:overview_imagenet}}, and then evaluate the validation accuracy of each model based on the \emph{image classes} in each partition. 
As Figure~\ref{fig:cause_gaia} shows, the validation accuracy is very consistent between the two sets of image classes when training the model in the IID setting: the results for IID DC-0 Model are shown, and IID DC-1 Model is the same. 
However, the validation accuracy varies drastically under the Non-IID setting (Non-IID DC-0 Model and Non-IID DC-1 Model). 
Specifically, both models perform well for the image classes in their respective partitions, but they perform very poorly for the image classes that are \emph{not} in their respective partitions.
This reveals that using \gaia in the Non-IID setting results in \emph{completely different} models among data partitions, and each model is only good for recognizing the image classes in its own data partition.

This raises the following question: How does \gaia produce completely different models in the Non-IID setting, given that \gaia synchronizes all significant updates ($\Delta w_j$) to ensure that the differences across models in each weight $w_j$ is insignificant (\xref{sec:background})? 
To answer this, we first compare each weight $w_j$ in the Non-IID DC-0 and DC-1 Models, and find that the average difference among all the weights is only 0.5\% (reflecting \pg{a 1\% threshold for significance in the last epoch}).
However, we find that given the same input image, the \emph{neuron} values are vastly different (with an average difference of 173\%). 
This finding suggests that small model differences can result in completely different models.
Mathematically, this is because weights can be positive or negative: a small percentage difference in individual weights can lead to a large percentage difference in the resulting neuron values, especially for neuron values that have small magnitudes.
As \gaia eliminates insignificant communication, it creates an opportunity for models in each data partition to specialize for the image classes in their respective data partition, at the expense of other classes.


\begin{figure}[h]
\centering
\includegraphics[width=0.55\textwidth]{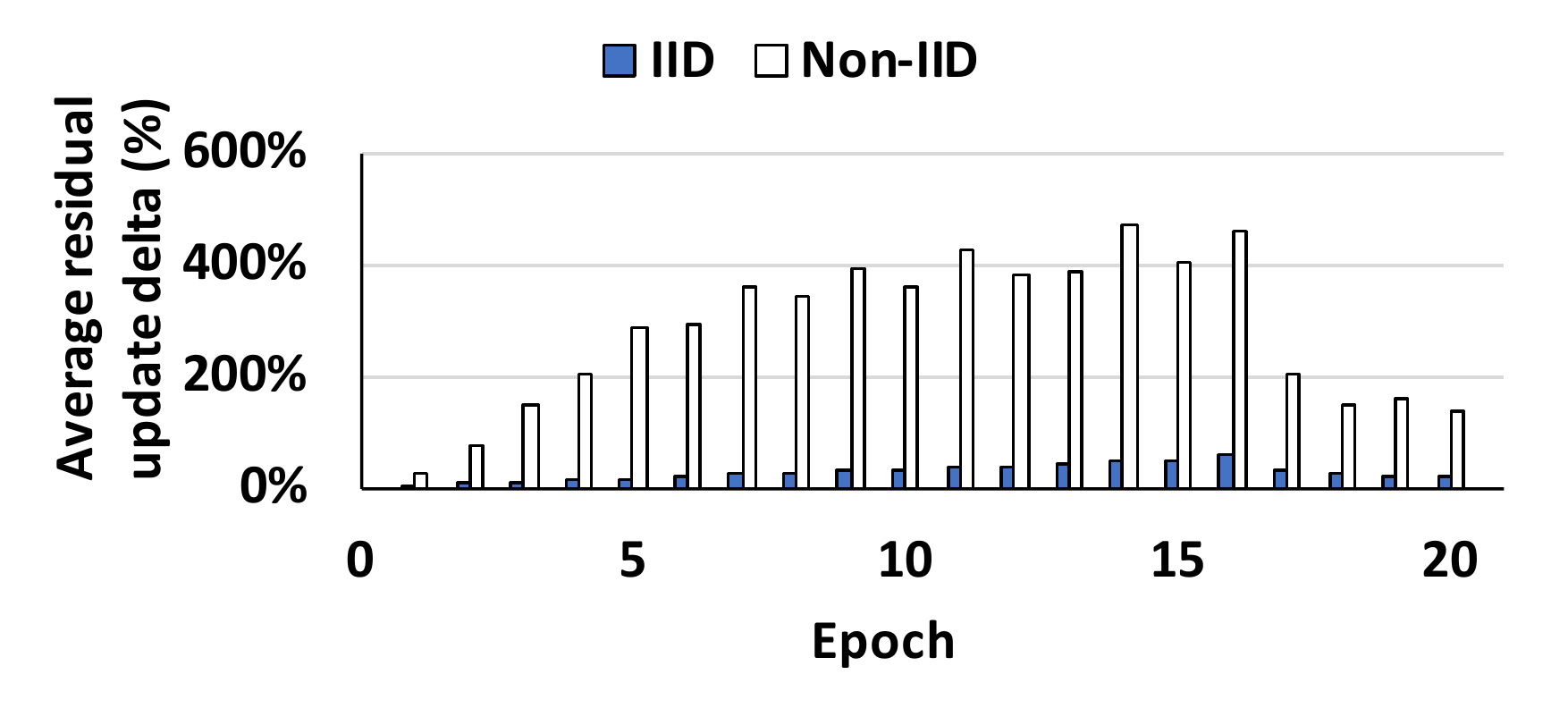}\vspace{-0.15in}
\caption{Average residual update delta (\%) for \dgc over the first 20 epochs.} 
\label{fig:update_delta_dgc}
\end{figure}

\textbf{DeepGradientCompression.}
\dgc and \fedavg always maintain \emph{one} global model, and hence there must be a \emph{different} reason for their model quality loss. 
For \dgc, we examine the average residual update delta ($||\Delta w_i / w_i||$). 
This number represents the magnitude of the gradients that have \emph{not} yet been exchanged among different $P_k$, as the algorithm communicates only a fixed number of gradients in each epoch (\xref{sec:background}). 
Thus, it can be viewed as the amount of gradient divergence among different $P_k$.
Figure~\ref{fig:update_delta_dgc} depicts the average residual update delta for the first 20 training epochs when training ResNet20 over CIFAR-10. 
(We show only the first 20 epochs because, \pg{as shown in Figure~\ref{fig:cifar10_resnet_curve}(d)}, training diverges after 20 epochs in the Non-IID setting.)
As the figure shows, the average residual update delta is an order of magnitude higher in the Non-IID setting (283\%) than that in the IID setting (27\%). 
Hence, each $P_k$ generates large gradients in the Non-IID setting, which is not surprising as each $P_k$ sees vastly different training data. 
However, these large gradients are not synchronized because \dgc sparsifies the gradients at a fixed rate. 
When they are finally synchronized, they may have diverged so much from the global model that they lead to the divergence of the whole model, and indeed our experiments \pg{often show such divergence.}

\begin{figure}[h]
\centering
\includegraphics[width=0.55\textwidth]{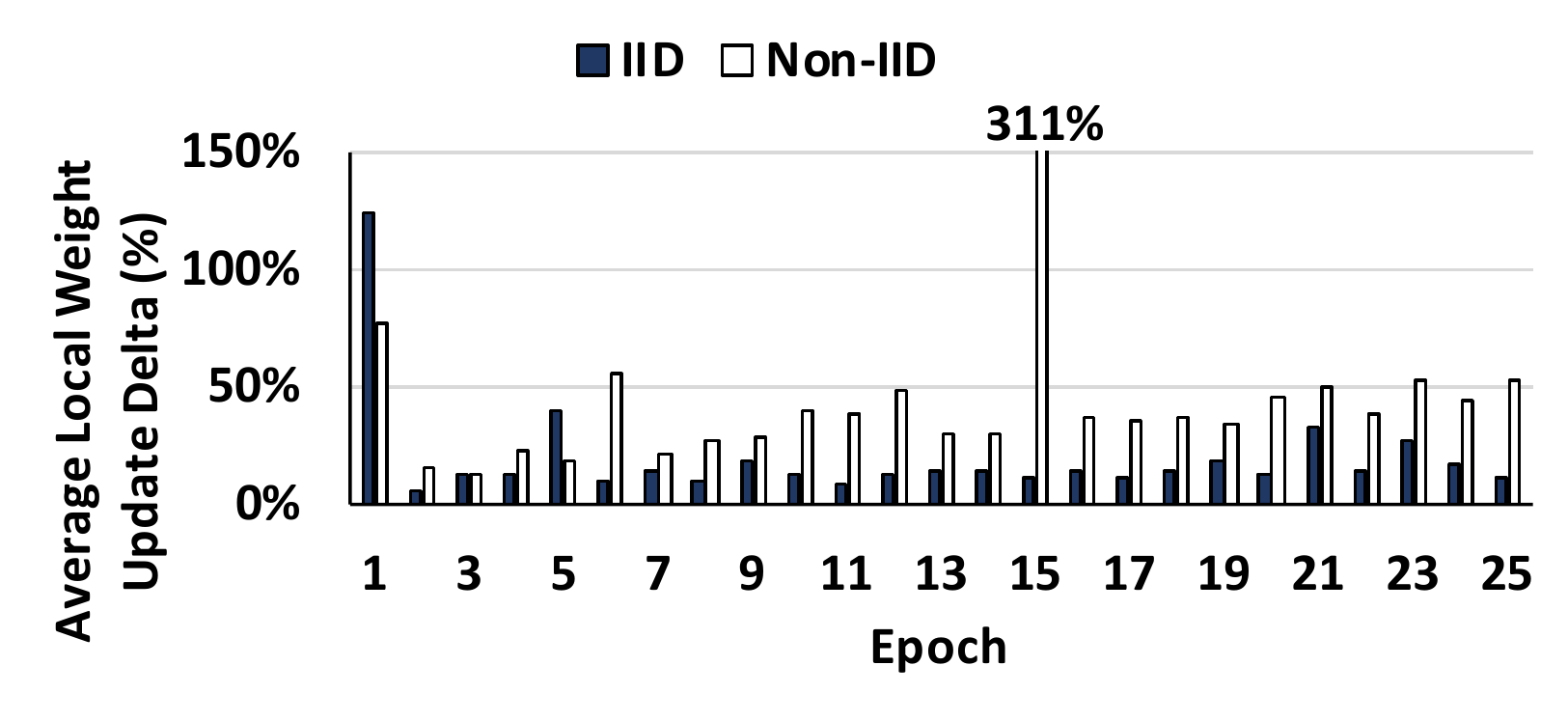}
\vspace{-0.15in}
\caption{Average local update delta (\%) for \fedavg over the first 25 epochs.} 
\label{fig:update_delta_fedavg}
\end{figure}

\textbf{FederatedAveraging.}
The analysis for \dgc can also apply to \fedavg, which delays communication from each $P_k$ by a fixed number of local iterations.
If the weights in different $P_k$ diverge too much, the synchronized global model can lose accuracy or completely diverge~\cite{DBLP:journals/corr/abs-1806-00582}. 
We validate this by plotting the average local weight update delta for \fedavg at each global synchronization point ($||\Delta w_i / w_i||$, where $w_i$ is the averaged global model weight). 
Figure~\ref{fig:update_delta_fedavg} depicts this number for the first 25 training epochs when training AlexNet over the CIFAR-10 dataset \pg{(Figure~\ref{fig:cifar10_alexnet_curve}(c)).}
As the figure shows, the average local weight update delta in the Non-IID setting (48.5\%) is much higher than that in the IID setting (20.2\%), which explains why Non-IID data partitions lead to major accuracy loss for \fedavg.
The difference is less pronounced than with \dgc, so the impact on accuracy is smaller with \fedavg.

\section{Details on Algorithm Hyper-Parameters}
\label{sec:decentral_parameter}

We study the sensitivity of the non-IID problem to hyper-parameter choice.
Tables \ref{tbl:gaia_hyper_parameter}, \ref{tbl:fedavg_hyper_parameter} and \ref{tbl:dgc_hyper_parameter} present the results for \gaia, \fedavg and \dgc, \pg{respectively,} by varying their respective hyper-parameters when training on CIFAR-10.
We compare the results with BSP. Two major observations are in order.

First, almost all hyper-parameter settings lead to significant accuracy loss in the Non-IID setting (relative to BSP in the IID setting).
Even with a relatively conservative hyper-parameter setting (e.g., $T_0 = 2\%$ for \gaia or $Iter_{Local} = 5$ for \fedavg, the most communication-intensive of the choices shown), we still observe a 3.3\% to 42.3\% accuracy loss. 
On the other hand, the exact same hyper-parameter choice in the IID setting can mostly achieve BSP-level accuracy (except for ResNet20, which is troubled by the batch normalization problem, \xref{sec:batch_norm}). 
We see the same trend with much more aggressive hyper-parameter settings as well (e.g., $T_0 = 40\%$ for \gaia). 
This shows that the problem of Non-IID data partitions is not specific to particular hyper-parameter settings, and that hyper-parameter settings that work well in the IID setting may perform poorly in the Non-IID setting.

Second, more conservative hyper-parameter settings (which implies more frequent communication among the $P_k$) often greatly decrease the accuracy loss in the Non-IID setting. 
For example, the validation accuracy with $T_0 = 2\%$ is significantly higher than the one with $T_0 = 30\%$ for \gaia. 
\pg{This supports \name's approach (\xref{sec:solution}) that more frequent communication among the $P_k$ leads to} higher model quality in the Non-IID setting (mitigating the ``tug-of-war'' among the $P_k$  (\xref{subsec:decentral})). 

\begin{table}[t]
  \centering
  \small
  \begin{tabular}{c cc cc cc cc}
    \toprule
    \multirow{2}{*}{Configuration} & \multicolumn{2}{c}{AlexNet} & \multicolumn{2}{c}{GoogLeNet} & \multicolumn{2}{c}{LeNet} & \multicolumn{2}{c}{ResNet20} \\
    \cmidrule(lr){2-3} \cmidrule(lr){4-5} \cmidrule(lr){6-7} \cmidrule(lr){8-9}
    & {IID} & {Non-IID} & {IID} & {Non-IID} & {IID} & {Non-IID} & {IID} & {Non-IID} \\ \midrule
    BSP &  74.9\% & 75.0\%	& 79.1\% & 78.9\% & 77.4\% & 76.6\% & 83.7\% & \bad{44.3\%} \\ \midrule
    $T_0 = 2\%$ & 73.8\% & \bad{70.5\%} & 78.4\% & \bad{56.5\%} & 76.9\% & \bad{52.6\%} & 83.1\% & \bad{48.0\%} \\ \midrule
    $T_0 = 5\%$ & 73.2\% & \bad{71.4\%} & 77.6\% & \bad{75.6\%} & \bad{74.6\%} & \bad{10.0\%} & 83.2\% & \bad{43.1\%} \\ \midrule
    $T_0 = 10\%$ & 73.0\% & \bad{10.0\%} & 78.4\% & \bad{68.0\%} & 76.7\% & \bad{10.0\%} & 84.0\% & \bad{45.1\%} \\ \midrule
    $T_0 = 20\%$ & \bad{72.5\%} & \bad{37.6\%} & 77.7\% & \bad{67.0\%} & 77.7\% & \bad{10.0\%} & 83.6\% & \bad{38.9\%} \\ \midrule
    $T_0 = 30\%$ & \bad{72.4\%} & \bad{26.0\%} & 77.5\% & \bad{23.9\%} & 78.6\% & \bad{10.0\%} & 81.3\% & \bad{39.4\%} \\ \midrule
    $T_0 = 40\%$ & \bad{71.4\%} & \bad{20.1\%} & 77.2\% & \bad{33.4\%} & 78.3\% & \bad{10.1\%} & 82.1\% & \bad{28.5\%} \\ \midrule
    $T_0 = 50\%$ & \bad{10.0\%} & \bad{22.2\%} & \bad{76.2\%} & \bad{26.7\%} & 78.0\% & \bad{10.0\%} & \bad{77.3\%} & \bad{28.4\%}  \\     
    \bottomrule \\
  \end{tabular}
  \vspace{-0.1in}
  \caption{\pg{CIFAR-10 Top-1 validation accuracy} varying \gaia's $T_0$ hyper-parameter.
  \pg{Configurations with more than 2\% accuracy loss relative to} BSP in the IID setting are highlighted in purple.  
  Note that larger settings for $T_0$ indicate larger communication savings.}
  \label{tbl:gaia_hyper_parameter}
\vspace{0.1in}
\end{table}

\ignore{
\begin{table*}[h!]
  \centering
  \small
  \begin{tabular}{c cc cc}
    \toprule
    \multirow{2}{*}{Configuration} & \multicolumn{2}{c}{LeNet} & \multicolumn{2}{c}{ResNet20} \\
    \cmidrule(lr){2-3} \cmidrule(lr){4-5} 
    & {IID} & {Non-IID} & {IID} & {Non-IID} \\ \midrule
    BSP &  77.4\% & 76.6\% & 83.7\% & \bad{44.3\%} \\ \midrule
    $T_0 = 2\%$ & 76.9\% & \bad{52.6\%} & 83.1\% & \bad{48.0\%} \\ \midrule
    $T_0 = 5\%$ & \bad{74.6\%} & \bad{10.0\%} & 83.2\% & \bad{43.1\%} \\ \midrule
    $T_0 = 10\%$ & 76.7\% & \bad{10.0\%} & 84.0\% & \bad{45.1\%} \\ \midrule
    $T_0 = 20\%$ & 77.7\% & \bad{10.0\%} & 83.6\% & \bad{38.9\%} \\ \midrule
    $T_0 = 30\%$ & 78.6\% & \bad{10.0\%} & 81.3\% & \bad{39.4\%} \\ \midrule
    $T_0 = 40\%$ & 78.3\% & \bad{10.1\%} & 82.1\% & \bad{28.5\%} \\ \midrule
    $T_0 = 50\%$ & 78.0\% & \bad{10.0\%} & \bad{77.3\%} & \bad{28.4\%}  \\     
    \bottomrule \\
  \end{tabular}
  \caption{Top-1 validation accuracy (CIFAR-10) varying \gaia's $T_0$ hyper-parameter for LeNet and ResNet20.
  The configurations with more than 2\% accuracy loss from BSP in the IID setting are highlighted.  
  Note that larger settings for $T_0$ indicate larger communication savings.}
  \label{tbl:gaia_hyper_parameter_ext}
\end{table*}
}

\begin{table}[h!]
  \centering
  \small
  \begin{tabular}{c cc cc cc cc}
    \toprule
    \multirow{2}{*}{Configuration} & \multicolumn{2}{c}{AlexNet} & \multicolumn{2}{c}{GoogLeNet} & \multicolumn{2}{c}{LeNet} & \multicolumn{2}{c}{ResNet20} \\
    \cmidrule(lr){2-3} \cmidrule(lr){4-5} \cmidrule(lr){6-7} \cmidrule(lr){8-9}
    & {IID} & {Non-IID} & {IID} & {Non-IID} & {IID} & {Non-IID} & {IID} & {Non-IID} \\ \midrule
    BSP &  74.9\% & 75.0\%	& 79.1\% & 78.9\% & 77.4\% & 76.6\% & 83.7\% & \bad{44.3\%} \\ \midrule
    $Iter_{Local} = 5$ & 73.7\% & \bad{62.8\%} & \bad{75.8\%} & \bad{68.9\%} & 79.7\% & \bad{67.3\%} & \bad{73.6\%} & \bad{31.3\%} \\ \midrule
    $Iter_{Local} = 10$ & 73.5\% & \bad{60.1\%} & \bad{76.4\%} & \bad{64.8\%} & 79.3\% & \bad{63.2\%} & \bad{73.4\%} & \bad{28.0\%} \\ \midrule
    $Iter_{Local} = 20$ & 73.4\% & \bad{59.4\%} & \bad{76.3\%} & \bad{64.0\%} & 79.1\% & \bad{10.1\%} & \bad{73.8\%} & \bad{28.1\%} \\ \midrule
    $Iter_{Local} = 50$ & 73.5\% & \bad{56.3\%} & \bad{75.9\%} & \bad{59.6\%} & 79.2\% & \bad{55.6\%} & \bad{74.0\%} & \bad{26.3\%} \\ \midrule
    $Iter_{Local} = 200$ & 73.7\% & \bad{53.2\%} & \bad{76.8\%} & \bad{52.9\%} & 79.4\% & \bad{54.2\%} & \bad{75.7\%} & \bad{27.3\%} \\ \midrule
    $Iter_{Local} = 500$ & 73.0\% & \bad{24.0\%} & \bad{76.8\%} & \bad{20.8\%} & 79.6\% & \bad{19.4\%} & \bad{74.1\%} & \bad{24.0\%} \\ \midrule
    $Iter_{Local} = 1000$ & 73.4\% & \bad{23.9\%} & \bad{76.1\%} & \bad{20.9\%} & 78.3\% & \bad{19.0\%} & \bad{74.3\%} & \bad{22.8\%}  \\     
    \bottomrule \\
  \end{tabular}
    \vspace{-0.1in}
  \caption{CIFAR-10 Top-1 validation accuracy \pg{varying \fedavg's $Iter_{Local}$ hyper-parameter. 
  Configurations with more than 2\% accuracy loss relative to BSP in the IID setting} are highlighted in purple.  
  Note that larger settings for $Iter_{Local}$ indicate larger communication savings.}
  \label{tbl:fedavg_hyper_parameter}
  \vspace{0.1in}
\end{table}

\begin{table}[h!]
  \centering
  \small
  \begin{tabular}{c cc cc cc cc}
    \toprule
    \multirow{2}{*}{Configuration} & \multicolumn{2}{c}{AlexNet} & \multicolumn{2}{c}{GoogLeNet} & \multicolumn{2}{c}{LeNet} & \multicolumn{2}{c}{ResNet20} \\
    \cmidrule(lr){2-3} \cmidrule(lr){4-5} \cmidrule(lr){6-7} \cmidrule(lr){8-9}
    & {IID} & {Non-IID} & {IID} & {Non-IID} & {IID} & {Non-IID} & {IID} & {Non-IID} \\ \midrule
    BSP &  74.9\% & 75.0\%	& 79.1\% & 78.9\% & 77.4\% & 76.6\% & 83.7\% & \bad{44.3\%} \\ \midrule
    $E_{warm} = 8$ & 75.5\% & \bad{72.3\%} & 78.3\% & \bad{10.0\%} & 80.3\% & \bad{47.2\%} & \bad{10.0\%} & \bad{10.0\%} \\ \midrule
    $E_{warm} = 4$ & 75.5\% & 75.7\% & 79.4\% & \bad{61.6\%} & \bad{10.0\%} & \bad{47.3\%} & \bad{10.0\%} & \bad{10.0\%} \\ \midrule
    $E_{warm} = 3$ & 75.9\% & 74.9\% & 78.9\% & \bad{75.7\%} & \bad{64.9\%} & \bad{50.5\%} & \bad{10.0\%} & \bad{10.0\%} \\ \midrule
    $E_{warm} = 2$ & 75.7\% & 76.7\% & 79.0\% & \bad{58.7\%} & \bad{10.0\%} & \bad{47.5\%} & \bad{10.0\%} & \bad{10.0\%} \\ \midrule
    $E_{warm} = 1$ & 75.4\% & 77.9\% & 78.6\% & \bad{74.7\%} & \bad{10.0\%} & \bad{39.9\%} & \bad{10.0\%} & \bad{10.0\%}  \\ 
    \bottomrule \\
  \end{tabular}
  \vspace{-0.1in}
  \caption{CIFAR-10 Top-1 validation accuracy \pg{varying \dgc's $E_{warm}$ hyper-parameter. 
  Configurations with more than 2\% accuracy loss relative to BSP in the IID setting} are highlighted in purple.  
  Note that smaller settings for $E_{warm}$ indicate larger communication savings.}
  \label{tbl:dgc_hyper_parameter}
\end{table}

\section{More Alternatives to Batch Normalization}
\label{appendix:batchnorm}

\textbf{Weight Normalization~\cite{DBLP:conf/nips/SalimansK16}.} 
Weight Normalization (WeightNorm) normalizes the weights in a DNN as opposed to the neurons (which is what BatchNorm and most other normalization techniques do). 
WeightNorm is not dependent on minibatches as it normalizes the weights. 
However, while WeightNorm can effectively control the variance of the neurons, it still needs a mean-only BatchNorm in many cases to achieve the model quality and training speeds of BatchNorm~\cite{DBLP:conf/nips/SalimansK16}. 
This mean-only BatchNorm makes WeightNorm vulnerable to the Non-IID setting again, because there is a large divergence in $\mu_{\mathcal{B}}$ among the $P_k$ in the Non-IID setting (\xref{subsec:batch_norm_problem}).

\textbf{Layer Normalization~\cite{DBLP:journals/corr/BaKH16}.} 
Layer Normalization (LayerNorm) is a technique that is inspired by BatchNorm. 
Instead of computing the mean and variance of a minibatch for each \emph{channel}, LayerNorm computes the mean and variance across all channels for each \emph{sample}. 
Specifically, if the inputs are four-dimensional vectors $\mathcal{B}\times\mathcal{C}\times\mathcal{W}\times\mathcal{H}$ (batch $\times$ channel $\times$ width $\times$ height), BatchNorm produces $\mathcal{C}$ means and variances along the $\mathcal{B} \times \mathcal{W} \times \mathcal{H}$ dimensions. 
In contrast, LayerNorm produces $\mathcal{B}$ means and variances along the $\mathcal{C} \times \mathcal{W} \times \mathcal{H}$ dimensions (per-sample mean and variance). 
As the normalization is done on a per-sample basis, LayerNorm is not dependent on minibatches. 
However, LayerNorm makes a key assumption that all inputs make similar contributions to the final prediction, but this assumption does not hold for some models such as convolutional neural networks, where the activation of neurons should not be normalized with non-activated neurons. 
As a result, BatchNorm still outperforms LayerNorm for these models~\cite{DBLP:journals/corr/BaKH16}. 

\textbf{Batch Renormalization~\cite{DBLP:conf/nips/Ioffe17}.} 
Batch Renormalization (BatchReNorm) is an extension to BatchNorm that aims to alleviate the problem of small minibatches (or inaccurate minibatch mean, $\mu_{\mathcal{B}}$, and variance, $\sigma_{\mathcal{B}}$). 
BatchReNorm achieves this by incorporating the estimated global mean ($\mu$) and variance ($\sigma$) during \emph{training}, and introducing two hyper-parameters to contain the difference between ($\mu_{\mathcal{B}}$, $\sigma_{\mathcal{B}}$) and ($\mu$, $\sigma$).
These two hyper-parameters are gradually relaxed such that the earlier training phase is more like BatchNorm, and the later phase is more like BatchReNorm.

We evaluate BatchReNorm with BN-LeNet over CIFAR-10 to see if BatchReNorm can solve the problem of Non-IID data partitions. 
We replace all BatchNorm layers with BatchReNorm layers, and we carefully select the BatchReNorm hyper-parameters so that BatchReNorm achieves the highest validation accuracy in both the IID and Non-IID settings.
Table~\ref{tbl:batch_renorm_results} shows the Top-1 validation accuracy. 
We observe that while BatchNorm and BatchReNorm achieve similar accuracy in the IID setting, they both perform worse in the Non-IID setting. 
In particular, while BatchReNorm performs much better than BatchNorm in the Non-IID setting (75.3\% vs. 65.4\%), BatchReNorm still loses $\sim\!\!3\%$ accuracy compared to the IID setting. 
This is not surprising, because BatchReNorm still relies on minibatches to a certain degree, and prior work has shown that BatchReNorm's performance still degrades when the minibatch size is small~\cite{DBLP:conf/nips/Ioffe17}. 
Hence, BatchReNorm cannot completely solve the problem of Non-IID data partitions, which is a more challenging problem than small minibatches.

\begin{table}[h]
  \centering
  \small
  \begin{tabular}{cc cc}
  \toprule
  \multicolumn{2}{c}{BatchNorm} & \multicolumn{2}{c}{BatchReNorm} \\
  \cmidrule(lr){1-2} \cmidrule(lr){3-4} 
  {IID} & {Non-IID} & {IID} & {Non-IID} \\ \midrule
  78.8\% & \bad{65.4\%} & 78.1\% & \bad{75.3\%} \\
  \bottomrule
  \end{tabular}
  \caption{Top-1 validation accuracy (CIFAR-10) with BatchNorm and BatchReNorm for BN-LeNet, using BSP with $K=2$ partitions.}
  \label{tbl:batch_renorm_results}
\end{table}

\section{Accuracy Loss Details}
\label{appendix:accuracy_loss}

\begin{figure}[h!]
  \centering
  \begin{subfigure}[t]{0.8\linewidth}
    \centering
    \includegraphics[width=0.8\textwidth]{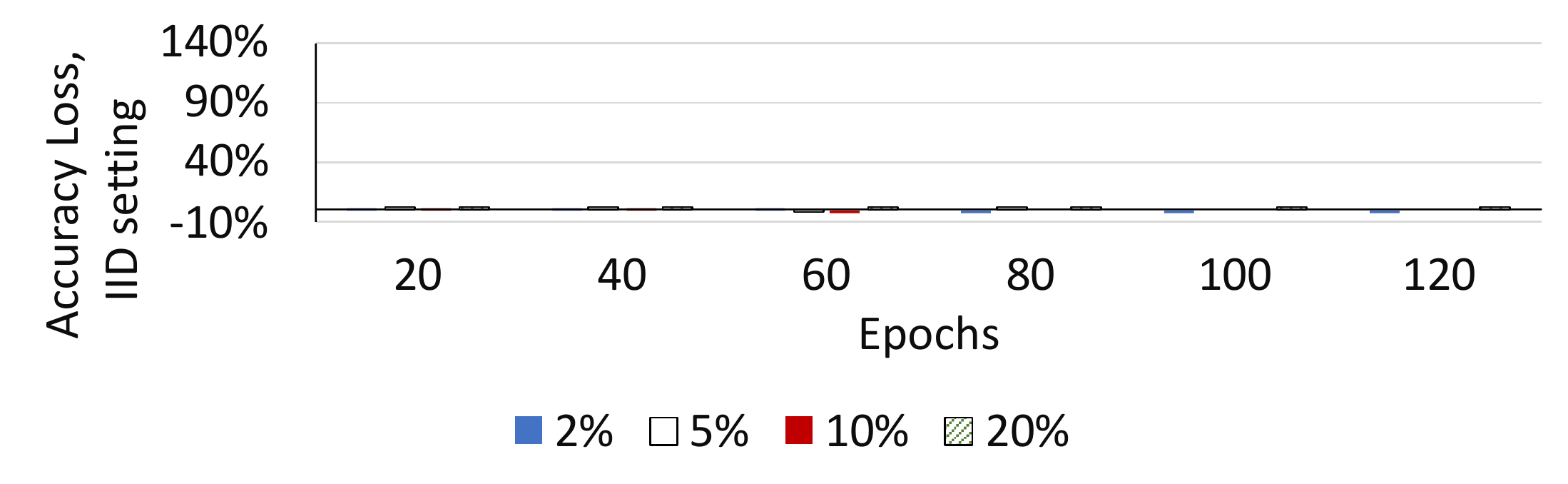}
  \end{subfigure}
   \centering 
  \begin{subfigure}[t]{0.8\linewidth}
    \centering
    \includegraphics[width=0.8\textwidth]{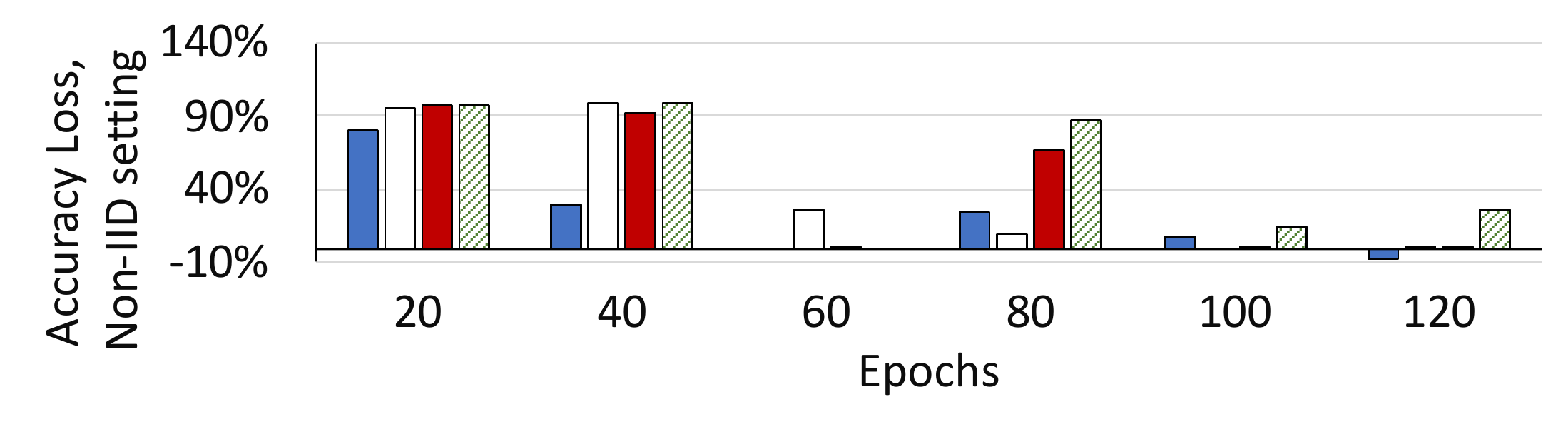}
  \end{subfigure}
  \vspace{-0.2in}
  \caption{Training accuracy loss over time (epochs) between data partitions when training \pg{GoogLeNet} over CIFAR-10 with \gaia. Each bar represents a $T_0$ for \gaia.}
  \label{fig:accuracy_drop}
\end{figure}

\pg{This section presents the full details of the findings summarized in \xref{subsec:comm_control}.}
Figure~\ref{fig:accuracy_drop} plots the accuracy loss between different data partitions when training \pg{GoogLeNet} over CIFAR-10 with \gaia. Two observations are in order. First, the accuracy loss changes drastically from the IID setting (0.4\% on average) to the Non-IID setting (39.6\% on average). This is expected as each data partition sees very different training data in the Non-IID setting, which leads to very different models in different data partitions. Second, more conservative hyper-parameters can lead to smaller accuracy losses in the Non-IID setting. For example, the accuracy loss for $T_0 = 2\%$ is significantly smaller than those for larger settings of $T_0$.
This is also intuitive as model divergence can be controlled by tightening communication between data partitions. 

\section{Discussion: Regimes of Non-IID Data}
\label{appendix:discussion}

Our study has focused on \emph{label-based} partitioning of data, in which the distribution of labels varies across partitions.
In this section, we present a broader taxonomy of regimes of non-IID data, as well as various possible strategies for dealing with non-IID data, the study of which we leave to future work.
We assume a general setting in which there may be many disjoint partitions, with each partition holding data collected from devices (mobile phones, video cameras, etc.) from a particular geographic region and time window.

\textbf{Violations of Independence.}
Common ways in which data tend to deviate from being independently drawn from an overall distribution are:
\squishlist
    \item \emph{Intra-partition correlation:} If the data within a partition are processed in an insufficiently-random order, e.g., ordered by collection device and/or by time, then independence is violated.  
    For example, consecutive frames in a video are highly correlated, even if the camera is moving.
    \item \emph{Inter-partition correlation:} Devices sharing a common feature can have correlated data across partitions.  
    For example, neighboring geo-locations have the same diurnal effects (daylight, workday patterns), have correlated weather patterns (major storms), and can witness the same phenomena (eclipses).
\squishlistend

\textbf{Violations of Identicalness.}
Common ways in which data tend to deviate from being identically distributed are:
\squishlist
    \item \emph{Quantity skew:} Different partitions can hold vastly different amounts of data.  
    For example, some partitions may collect data from fewer devices or from devices that produce less data.
    \item \emph{Label distribution skew:} Because partitions are tied to particular geo-regions, the distribution of labels varies across partitions.
    For example, kangaroos are only in Australia or zoos, and a person's face is only in a small number of locations worldwide.
    The study in this paper focused on this setting.
    \item \emph{Same label, different features:} The same label can have very different ``feature vectors'' in different partitions, e.g., due to cultural differences, weather effects, standards of living, etc.
    For example, images of homes can vary dramatically around the world and items of clothing vary widely.  
    Even within the U.S., images of parked cars in the winter will be snow-covered only in certain parts of the country.  
    The same label can also look very different at different times, at different time scales: day vs.~night, seasonal effects, natural disasters, fashion and design trends, etc.
    \item \emph{Same features, different label:} Because of personal preferences, the same feature vectors in a training data item can have different labels.  
    For example, labels that reflect sentiment or next word predictors have personal/regional biases.
\squishlistend

As noted in some of the above examples, non-IID-ness can occur over both time (often called \emph{concept drift}) and space (geo-location).

\textbf{Strategies for dealing with non-IID data.}
The above taxonomy of the many regimes of non-IID data partitions naturally leads to the question of what should the objective function of the DNN model be.
In our study, we have focused on obtaining a global model that minimizes an objective function over the union of all the data.
An alternative objective function might instead include some notion of ``fairness'' among the partitions in the final accuracy on their local data~\cite{DBLP:conf/iclr/LiSBS20}.
There could also be different strategies for treating different non-IID regimes.

As noted in Section~\ref{sec:related}, multi-task learning approaches have been proposed for jointly training local models for each partition, but a global model is essential whenever a local model is unavailable or ineffective.
A hybrid approach would be to train a ``base'' global model that can be quickly ``specialized'' to local data via a modest amount of further training on local data~\cite{DBLP:journals/corr/abs-2002-04758}.
This approach would be useful for differences across space and time.
For example, a global model trained under normal circumstances could be quickly adapted to natural disaster settings such as hurricanes, flash floods and forest fires.

As one proceeds down the path towards more local/specialized models, it may make sense to cluster partitions that hold similar data, with one model for each cluster~\cite{DBLP:journals/corr/abs-2002-10619, DBLP:journals/corr/abs-2004-11791, DBLP:journals/corr/abs-2002-11223}.
The goal is to avoid a proliferation of too many models that must be trained, stored, and maintained over time.

Finally, another alternative for handling non-IID data partitions is to use multi-modal training that combines DNNs with key attributes about the data partition pertaining to its geo-location.
A challenge with this approach is determining what the attributes should be, in order to have an accurate yet reasonably compact model
(otherwise, in the extreme, the model could devolve into local models for each geo-location).

\end{document}